\documentclass{article}
\usepackage{iclr2027_conference,times}

\usepackage{microtype}
\usepackage{graphicx}
\usepackage{subcaption}
\usepackage{booktabs}
\usepackage{url}
\usepackage{hyperref}

\usepackage{amsmath}
\usepackage{amssymb}
\usepackage{mathtools}
\usepackage{amsthm}
\usepackage{algorithm}
\usepackage{algorithmic}
\usepackage[capitalize,noabbrev]{cleveref}

\theoremstyle{plain}

\theoremstyle{definition}

\newcommand{\opftloop}{OPFT loop}
\newcommand{\opft}{\textsc{RFPO}}

\iclrfinalcopy

\title{Ratio-Free Policy Optimization:\\A Bias-Variance View of On-Policy Fine-Tuning}

\author{Xiaohu Xie\thanks{Corresponding author: \texttt{xxie@amazon.com}}, Shahin Boluki, Ketan Ramaneti, Apoorva Kumar, Benjamin Z. Yao \\
Amazon \\
\texttt{\{xxie,shahinbo,kramanet,apookr,benjamy\}@amazon.com}}

\begin{document}

\maketitle
\lhead{Preprint. Under review.}

\begin{abstract}
Practical LLM reinforcement learning is never strictly on-policy: each batch
of rollouts is reused for many optimizer steps, and rollouts come from a separate
inference engine. We show that GRPO degrades under both drifts --- and on coding
collapses --- and explain why. We formalize \emph{on-policy fine-tuning (OPFT)},
updating a model on its own rollouts, as the \emph{\opftloop{}}, of which GRPO and
iterative rejection-sampling SFT are realizations, and propose \opft{} (Ratio-Free Policy Optimization), a novel realization
that optimizes the surrogate $J_{\text{RF}}(\pi_\theta) = \mathbb{E}_{\pi_{\text{old}}}
[A \log \pi_\theta]$ --- a reward-weighted maximum-likelihood objective
requiring no importance ratio --- as a local proxy for the on-policy
expected reward $J(\pi_\theta)$. We identify three mechanisms --- z-score advantages,
negative samples, and trust-region clipping --- that make this surrogate stable,
each with a distinct role verified by theory and ablations; the analysis also
explains the well-known collapse of iterative rejection-sampling SFT. GRPO realizes
analogous mechanisms through different machinery: the core updates of \opft{} and
GRPO coincide at strict on-policy and diverge under the practical sampling gaps that
any modern LLM RL stack faces (multiple optimizer steps per batch;
decoupled rollout/training engines). A bias-variance analysis, checked by
directly measuring both gradient estimators against a fresh on-policy gradient,
shows that the importance-weighted estimator's error grows with drift while the
ratio-free one pays a surrogate bias that a per-token trust region keeps small,
so \opft{} has the lower MSE once drift is large. On Qwen3-4B trained on POLARIS
(math) and DeepCoder (coding), we verify this along both drift axes. As optimizer
steps per batch grow from 1 to 8 to 32, GRPO degrades from 0.879 to 0.861 to 0.824
on MATH-500 and collapses on LiveCodeBench-v6 (0.67 to 0.38 to 0.24), while \opft{}
stays at 0.873--0.875 and 0.67, matching strict on-policy GRPO. Under bypass-mode
rollout drift (8 steps per batch), GRPO collapses ($0.861 \to 0.718$ on MATH-500,
$0.38 \to 0.04$ on coding) while \opft{} is nearly unaffected (0.866 / 0.67).
Needing no reference model, KL term or importance ratio, \opft{} is also
1.4$\times$ faster per step than KL-free GRPO at the same mini-batch size on math.
\end{abstract}

\section{Introduction}
\label{sec:intro}

Pre-training and supervised fine-tuning (SFT) fit a model by maximum likelihood
to a fixed external corpus; reinforcement learning (RL) post-training instead
trains the model on its \emph{own} rollouts, scored by a reward, so the training
distribution shifts as the policy updates. This on-policy property is what makes
RL effective --- the gradient points where the current policy can improve --- and
RL is often reported to forget less than continued SFT~\citep{shenfeld2025rl}.
GRPO~\citep{shao2024deepseekmath}, an actor-only group-relative policy-gradient
recipe, has driven much of the recent progress in LLM reasoning.

In this paper we formalize this practice --- training on the model's
own rollouts --- as the \emph{on-policy fine-tuning (OPFT)} loop, an abstract algorithm in
which generation-and-scoring alternates with mini-batch updates
(\cref{alg:opft_loop} in \cref{sec:loop}). GRPO is one realization of the \opftloop{}; the
natural maximum-likelihood realization, iterative rejection-sampling
SFT~\citep{gulcehre2023reinforced, zelikman2022star, yuan2023scaling},
is another. We ask two questions. First, what is the right
\emph{objective} for the update step inside the \opftloop{} --- and is
policy gradient with importance sampling actually necessary? Second,
given an objective, what \emph{mechanisms} are required to keep the
\opftloop{} stable? Our proposed answer is \opft{} (Ratio-Free Policy Optimization), a
recipe whose two contributions we summarize next.

\paragraph{Strict on-policy is impractical.}
We use ``on-policy'' in the standard PPO/GRPO sense: rollouts come from the
current policy and may be reused for several optimizer steps within a batch. The
cleanest setting --- rollout policy and actor identical at every gradient step ---
is inefficient and impractical for any modern
LLM RL stack, for two distinct reasons:
\begin{enumerate}
\item \textbf{Policy drift (within-batch), from training efficiency.} Strict
on-policy requires one optimizer step per training batch (mini-batch size
equal to batch size). Generating thousands of rollouts to take a single
gradient step wastes compute. Practical recipes use
\texttt{mbs < bs}, taking $K = \texttt{bs}/\texttt{mbs}$ optimizer steps per batch
(\texttt{mbs} counts prompts; this is sample reuse, not gradient accumulation,
which keeps $K=1$ and introduces no drift).
The actor $\pi_\theta$ then drifts from the batch-start snapshot
$\pi_{\text{old}}$ across these steps; we call this deviation
$\pi_\theta - \pi_{\text{old}}$ the \emph{policy drift}, and it grows with the
number of steps.
\item \textbf{Rollout drift (engine mismatch), from decoupled inference and training.}
Modern frameworks separate the rollout engine (vLLM~\citep{kwon2023efficient},
sglang) from the training engine (e.g., FSDP), and even at the same checkpoint their
numerics differ (attention kernels, precision, quantization). So the rollout policy $\pi^*_{\text{old}}$ --- the
distribution the samples were actually drawn from, realized by the inference
engine --- differs from the training-engine snapshot $\pi_{\text{old}}$. We
call this deviation $\pi_{\text{old}} - \pi^*_{\text{old}}$ the \emph{rollout
drift}; unlike policy drift it is present even at ``on-policy'' and is static
within a batch.
\end{enumerate}

\emph{Policy drift} and \emph{rollout drift} are universal --- every production
realization of the \opftloop{} faces them --- and they are the two axes along which
we compare \opft{} and GRPO empirically (\cref{sec:experiments}) and
analyze them theoretically (\cref{sec:method_biasvar}). The actor's total
deviation from the distribution its samples were drawn from decomposes as
$\pi_\theta - \pi^*_{\text{old}} = (\pi_\theta - \pi_{\text{old}}) +
(\pi_{\text{old}} - \pi^*_{\text{old}})$ --- policy drift plus rollout drift.
This is also why the bias-variance view is the right lens: at strict on-policy there
is no drift to trade off, but in practice it is unavoidable, and methods
differ in how they navigate it.

\paragraph{Contributions.}
We make three contributions.

\textbf{(1) A bias-variance view of the \opftloop{}.}
We frame GRPO and \opft{} as two strategies for estimating the same
on-policy gradient from off-policy rollouts: GRPO importance-corrects each
token (less surrogate bias, but the ratio $r_t = \pi_\theta /
\pi_{\text{old}}$ adds a second moment that grows with drift and that clipping
does not fully cap), while \opft{} optimizes a surrogate $J_{\text{RF}}$ whose
definition needs no importance correction (no ratio, but a surrogate bias that a
trust region controls locally). The two coincide at strict on-policy and diverge under drift.
Measuring both estimators directly (\cref{sec:method_biasvar}), we find that the
ratio-free one has up to 2.5$\times$ lower MSE once drift is large, while at small
drift the importance-weighted one is as good or better. In short:
under large drift, a controlled bias beats the importance-weighted estimator's growing error.

\textbf{(2) \opft{}: a novel recipe for the \opftloop{} with three
mechanisms.}
We propose \opft{} (Ratio-Free Policy Optimization), a concrete recipe that realizes
a surrogate objective with three stabilizing mechanisms: \emph{z-score advantages}
(M1), \emph{negative samples} via a centered advantage (M2), and a
per-token \emph{trust region} via absolute probability clipping (M3).
Ablations show M2 and M3 are necessary for stability and M1 for smooth
optimization, and none of six alternative stabilizers recovers stability.
Theoretically, the three play distinct roles --- M1 balances per-prompt
gradient contribution, M2 removes the degenerate fixed point of positive-only
refitting, M3 locally controls the bias of $J_{\text{RF}}$ relative to $J$. GRPO realizes the same three
mechanisms through different machinery (\cref{sec:method_mechanisms}).

\textbf{(3) A simpler, faster recipe.} \opft{} needs no reference model, KL term
or importance ratio; at \texttt{mbs}=8 it is 1.4$\times$ faster per step than
KL-free GRPO on math and about 2$\times$ faster than GRPO on coding, both at the same
\texttt{mbs} (\cref{sec:summary}),
and changes only the update step, so it is a drop-in replacement for GRPO's update
in existing pipelines.

We verify both contributions on Qwen3-4B across math (POLARIS-53K) and coding
(DeepCoder-16K): \opft{} matches strict-on-policy GRPO and holds under policy drift
(smaller \texttt{mbs}); under bypass-mode rollout drift at 8 steps per batch GRPO
collapses while \opft{} is nearly unaffected, and on coding \opft{} stays far ahead
even at 32 (\cref{sec:experiments}).

\section{Related Work}
\label{sec:related}

\paragraph{Policy-gradient and iterative-MLE realizations of the
\opftloop{}.}
The dominant realizations of the \opftloop{} for LLM post-training are
policy-gradient methods --- PPO \citep{schulman2017proximal} for RLHF
\citep{ouyang2022training}, GRPO \citep{shao2024deepseekmath} which we
compare against, DAPO \citep{yu2025dapo} --- all of which optimize a
clipped policy-gradient objective with importance sampling between
$\pi_\theta$ and $\pi_{\text{old}}$. Critic-free REINFORCE-style estimators such
as RLOO \citep{ahmadian2024back} are ratio-free but assume fresh on-policy
samples; \opft{}'s gradient has the same form but is applied to
\emph{reused} samples, and our contribution is the analysis of that reuse and the
mechanisms that keep it stable. Closest to \opft{} is LGPO
\citep{zhang2026is}, which also optimizes the advantage-weighted log-likelihood
without old-policy likelihoods. The two differ in three ways. (1) \emph{Trust
region:} LGPO zeroes whole sequences whose current average log-likelihood exceeds a
fixed threshold, regardless of the advantage sign and of any snapshot; \opft{} masks
individual tokens whose probability moved more than $\delta$ from the batch-start
snapshot in the direction of the advantage, which caps the drift that drives the
surrogate's bias. (2) \emph{Scope:} LGPO studies 1.5B--3B models with a few (2--4)
updates per rollout batch in its main experiments; we study up to 32 inner updates,
engine drift, and coding, where GRPO collapses. (3) \emph{Analysis:} we measure when
dropping the ratio pays off (\cref{sec:method_biasvar}). The natural maximum-likelihood
counterpart is iterative rejection-sampling SFT
\citep{gulcehre2023reinforced, zelikman2022star, yuan2023scaling}, which
is conceptually simpler but is widely reported to collapse iteratively
\citep{singh2024beyond}; our analysis identifies the missing ingredient
as the centered-advantage / negative-sample mechanism M2 of
\cref{sec:method}. DPO \citep{rafailov2023direct} provides a related
contrastive signal, anchored to a reference model. Most directly
comparable to our surrogate framing is MaxRL
\citep{tajwar2026maxrl}, which interpolates between standard RL and
exact maximum-likelihood estimation through a compute-indexed family of
sample-based objectives. Our work is complementary: rather than indexing by
sampling compute, we identify the practical drift sources that any
modern LLM RL stack faces (\cref{sec:intro}) and characterize the
bias-variance trade-off along those axes.

\paragraph{Stability under rollout drift.}
verl's \emph{bypass mode} reuses vLLM rollout log-probabilities as the IS
denominator, exposing the gap between the rollout policy $\pi^*_{\text{old}}$ and
the actor; we use it as an explicit drift axis (\cref{sec:drift_bypass}).
Concurrent works move the ratio or trust region from the token to the sequence or
prefix: GSPO \citep{zheng2025gspo} uses a single sequence-level ratio, MinPRO
\citep{lei2026minpro} the minimum token-level ratio over the preceding prefix, and
trust-region masking \citep{li2025trustregion} masks entire sequences that violate a
bound on the maximum token-level divergence. \opft{} sits at the per-token end of
this axis (M3); the surrogate $J_{\text{RF}}$ is independent of where the trust
region is placed, and a sequence-level \opft{} variant is left to future work
(\cref{app:future_work}). DAPO's non-ratio components (dynamic sampling, overlong
shaping) are likewise complementary.

\paragraph{Bias-variance trade-offs in off-policy RL.}
Trading IS variance for surrogate bias is a long-standing pattern --- a
core motivation of trust-region methods (TRPO, PPO) and conservative
policy iteration. Our contribution is to apply this lens to LLM
on-policy fine-tuning explicitly: characterizing GRPO and \opft{} as
two corners of the trade-off, and measuring directly when the surrogate's
gradient has the lower MSE (only once drift is large).

\section{The \opftloop{} and Its Realizations}
\label{sec:loop}

We study the abstract \opftloop{} in \cref{alg:opft_loop}: at each
training batch, the policy generates and scores rollouts on a set of
prompts, snapshots itself, and then performs
$K = \texttt{bs} / \texttt{mbs}$ optimizer steps over
mini-batches of the same rollout set before drawing the next batch.
The \opftloop{} is general enough to describe every method studied in
this paper --- realizations differ only in the implementation of
$\textsc{Update}$ inside the inner mini-batch loop; the outer
rollout-and-score loop is shared.

\begin{algorithm}[t]
\caption{The OPFT loop (abstract)}
\label{alg:opft_loop}
\begin{algorithmic}[1]
\REQUIRE Policy $\pi_\theta$, prompts $\{x_i\}$, reward $R$,
         rollouts per prompt $N$,
         batch size \texttt{bs}, mini-batch size \texttt{mbs},
         update routine $\textsc{Update}$
\FOR{each training batch of size \texttt{bs}}
  \STATE $\pi_{\text{old}} \gets \pi_\theta$ \hfill \emph{(snapshot)}
  \FOR{each prompt $x_i$ in the batch}
    \STATE $Y_i \gets \{y_{i,1}, \dots, y_{i,N}\}
            \sim \pi^*_{\text{old}}(\cdot \mid x_i)$
            \hfill \emph{(rollouts)}
    \STATE $r_{i,j} \gets R(x_i, y_{i,j})$
            \hfill \emph{(scoring)}
  \ENDFOR
  \FOR{each mini-batch $m$ of \texttt{mbs} prompt groups in $(X, Y, R)$}
    \STATE $\pi_\theta \gets \textsc{Update}(\pi_\theta,\, \pi_{\text{old}},\, m)$
            \hfill \emph{(design choice)}
  \ENDFOR
\ENDFOR
\end{algorithmic}
\end{algorithm}

\textbf{Static-reference methods} train against a rollout distribution
that does not respond to the model: pretraining and SFT (outside the loop
altogether), and single-round rejection-sampling SFT
\citep{gulcehre2023reinforced, zelikman2022star, yuan2023scaling}, the loop's
\emph{frozen boundary} (one outer round, \texttt{bs} $\geq$ dataset size). Both are
stable because the training distribution does not respond to the model.

\textbf{Drifting-reference methods} re-snapshot the rollout policy each
outer round (\texttt{bs} $<$ dataset size), so $\pi_{\text{old}}$ tracks
$\pi_\theta$ and the rollout distribution drifts with training. The natural
maximum-likelihood realization, \emph{iterative} RS-SFT, regenerates
rollouts from the updated policy each round and trains MLE on the
keepers --- and is widely reported to collapse \citep{singh2024beyond}.
GRPO \citep{shao2024deepseekmath} is the canonical stable realization,
combining a clipped policy-gradient objective with importance sampling,
group-relative z-score advantages, and a KL term against the
initialization. Our recipe \opft{} (\cref{sec:method}) is a third
drifting-reference realization.

Within these realizations, both drift axes of \cref{sec:intro} live in the inner
loop: policy drift ($\pi_\theta \neq \pi_{\text{old}}$) and rollout drift
($\pi^*_{\text{old}} \neq \pi_{\text{old}}$), where $\pi_{\text{old}}$ is the
training-engine snapshot used as the trust-region anchor and, for GRPO, the IS
denominator.

Iterative RS-SFT is the one that collapses; \cref{sec:method_objective} shows it
is the \emph{naive} \opft{} recipe.

\section{\opft{}: A Stable Realization of the \opftloop{}}
\label{sec:method}

This section develops \opft{}, our recipe for the \opftloop{}. We begin
with the objective (\cref{sec:method_objective}), analyze it through the
bias-variance lens against GRPO's importance-sampling alternative
(\cref{sec:method_biasvar}), then introduce three mechanisms grounded in this
analysis (\cref{sec:method_mechanisms}); \cref{sec:method_algorithm} states the
full update routine.

\subsection{A ratio-free objective $J_{\text{RF}}$}
\label{sec:method_objective}

\emph{Throughout this section and \cref{sec:method_biasvar} we assume
no rollout drift}, i.e., $\pi^* = \pi$; we restore
rollout drift at the end of \cref{sec:method_biasvar}. 
The on-policy expected reward $J(\pi_\theta) = \mathbb{E}_{y \sim \pi_\theta}[R(y)]$ is the quantity we ultimately want to maximize. 
The gradient of $J$ is an
expectation under the \emph{current} policy $\pi_\theta$ while the rollouts are
sampled under $\pi_{\text{old}}$. GRPO closes the gap with an importance-sampling
(IS) correction $r_t = \pi_\theta(y_t|x,y_{<t}) / \pi_{\text{old}}(y_t|x,y_{<t})$,
yielding the IS surrogate $J_{\text{IS}}(\pi_\theta) \;=\; \mathbb{E}_{y \sim
\pi_{\text{old}}} \bigl[\sum_t r_t\, A(y)\bigr]$ (before clipping), whose gradient is a per-token off-policy estimate of $\nabla J$: it
corrects each token's conditional but not the distribution of its prefix
$y_{<t}$ or of the continuation that determines $A$, so it is exact only at
strict on-policy. We propose instead to change the local target to
the surrogate
\begin{equation}
J_{\text{RF}}(\pi_\theta) \;=\; \mathbb{E}_{y \sim \pi_{\text{old}}}\bigl[A(y) \log
\pi_\theta(y)\bigr],
\label{eq:opft_surrogate}
\end{equation}
whose expectation is taken under the sampling distribution itself, so
no importance ratio is required. At $\theta = \theta_{\text{old}}$,
$\nabla J_{\text{RF}} = \nabla J$; a small step on $J_{\text{RF}}$ improves $J$ to first order. Due to autoregressive factorization, $J_{\text{RF}}$'s gradient decomposes into a sum of per-token score-function gradients, weighted by the sequence-level advantage $A(y)$. Similar to GRPO, RFPO operates on this token-level decomposition.

\paragraph{Connection to RS-SFT, and an MLE implementation.}
Iterative rejection-sampling SFT
\citep{gulcehre2023reinforced, zelikman2022star, yuan2023scaling}
is a special case of optimizing $J_{\text{RF}}$ with the binary advantage $A(y) = \mathbf{1}[R(y) = 1]$, i.e.,
train MLE on the keepers, ignore the rest. 
More generally, $J_{\text{RF}}$ is a reward-weighted maximum-likelihood
objective in the family of reward- and advantage-weighted regression
\citep{peters2007rwr, peng2019awr}; our contribution is not this
objective but the ratio-free bias-variance analysis
(\cref{sec:method_biasvar}) and the mechanisms that stabilize it under
drift. Because
$J_{\text{RF}}$ has no importance ratio, any choice of $A$ admits an iterative
weighted-MLE implementation: snapshot $\pi_{\text{old}}$, generate
rollouts, compute $A$, and minimize $-\sum A(y) \log \pi_\theta(y)$
over mini-batches.

\subsection{Bias-variance trade-off}
\label{sec:method_biasvar}

Both \opft{} and GRPO target $\nabla J(\pi_\theta) =
\mathbb{E}_{y \sim \pi_\theta}[A \nabla \log \pi_\theta(y)]$ from rollouts
$y \sim \pi_{\text{old}}$ (no rollout drift, per \cref{sec:method_objective}; we
restore it below). We compare their per-sample token gradient estimators $\hat g$ through
$\mathrm{MSE}(\hat g) = \mathrm{Var}(\hat g) + \|\mathrm{Bias}(\hat
g)\|^2$, the error of $\hat g$ as an estimate of $\nabla J$. Off-policy samples force a choice in how to spend this error:
correct the sampling mismatch with an importance ratio, \emph{reducing
the bias} but paying the ratio's extra second moment, which grows with policy drift; or change the target
so its expectation already matches the sampler, \emph{removing the ratio} but paying a
surrogate bias. GRPO and \opft{} sit at these two corners.

\paragraph{GRPO --- correct the ratio.}
Reweighting by $r_t = \pi_\theta/\pi_{\text{old}}$ gives
$\hat g_{\text{IS}} = r_t \cdot A \cdot \nabla \log \pi_\theta$, the
per-sample gradient of $J_{\text{IS}}$ (\cref{sec:method_objective}).
Up to clipping, its bias is the residual prefix and continuation mismatch ---
accepted to avoid the instability of the exact, cumulative prefix ratio
\citep{lei2026minpro}. Its second moment $\mathbb{E}_{\pi_{\text{old}}}[r_t^2
\|A\nabla\log\pi_\theta\|^2]$ carries the ratio's extra factor, which is $1$ at
strict on-policy and grows with policy drift. Ratio clipping does not remove it:
it zeroes a token's gradient only once $r_t$ leaves $[1-\epsilon, 1+\epsilon]$ in
the advantage-improving direction, so a negative-advantage token whose probability
rose keeps $r_t A \nabla\log\pi_\theta$ with $r_t > 1+\epsilon$, capped only by
verl's dual clip (\cref{app:theory_proofs}). The effect compounds across the
$K = \texttt{bs}/\texttt{mbs}$ optimizer steps in a batch: a token pushed up at
step~1 has $r_t > 1$ at step~2, amplifying the next update.

\paragraph{\opft{} --- drop the ratio.}
\opft{} instead optimizes the surrogate $J_{\text{RF}}$~\eqref{eq:opft_surrogate},
whose expectation already matches the sampler, so its per-sample gradient
$\hat g_{\text{RF}} = A \cdot \nabla \log \pi_\theta$ carries no ratio. It is
therefore unbiased for $\nabla J_{\text{RF}}$ but biased for $\nabla J$ --- the price of dropping the ratio.
This bias is governed by the policy drift $\Delta(y) = \pi_\theta(y) - \pi_{\text{old}}(y)$:
\begin{equation}
\nabla J - \nabla J_{\text{RF}} = \textstyle\sum_y \Delta(y)\, A(y)\, \nabla \log \pi_\theta(y),
\label{eq:rfpo_bias}
\end{equation}
which is zero at strict on-policy and grows with the drift. M3 controls it \emph{locally}, capping each token's drift
in the advantage direction at $\delta$ (\cref{app:theory_proofs}).
Its variance has no ratio factor, but it is not independent of drift either, since
$\nabla\log\pi_\theta$ is evaluated at the drifting $\theta$. Neither expression
settles the comparison: which estimator has the lower MSE depends on how fast the
importance-weighted estimator's error grows relative to the surrogate bias. We
therefore measure it.

\begin{figure}[t]
\centering
\includegraphics[width=\columnwidth]{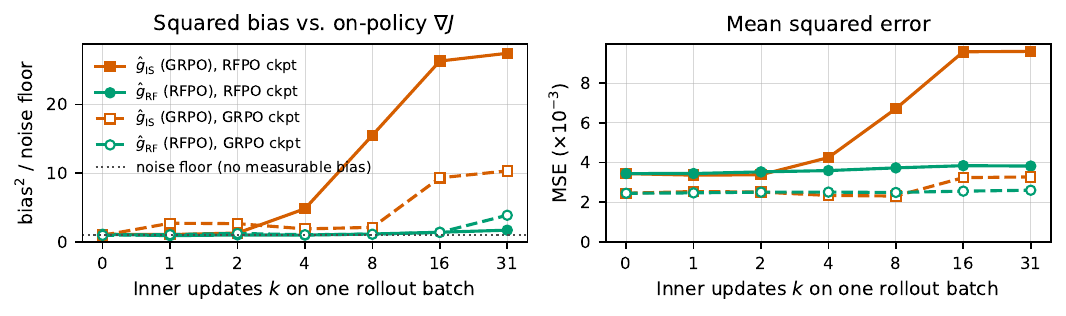}
\caption{Gradient-estimator probe. From a step-25 checkpoint we take one rollout
batch, run $k$ inner updates on it, and compare $\hat g_{\text{IS}}$ (GRPO) and
$\hat g_{\text{RF}}$ (\opft{}) with a fresh on-policy estimate of $\nabla J$.
Left: squared bias relative to the noise floor ($\approx$1 means no measurable bias).
Right: MSE. Filled markers and solid lines: checkpoint of an \opft{} run, \opft{}
inner updates; open markers and dashed lines: checkpoint of a GRPO run, GRPO inner
updates with its own optimizer state. Protocol in \cref{app:gradprobe}.}
\label{fig:gradprobe}
\end{figure}

\paragraph{Direct measurement.}
From a training checkpoint we take one rollout batch (256 prompts $\times$ 16
rollouts), run $k = 0, \dots, 31$ inner updates on it in mini-batches of 8 prompts,
and at each recorded $k$ evaluate both estimators on every mini-batch of the batch
against a fresh on-policy estimate of $\nabla J$ drawn from the updated policy
(\cref{fig:gradprobe}; protocol in \cref{app:gradprobe}). At the step-25
checkpoint of an \opft{} run, the squared bias of $\hat g_{\text{IS}}$ grows to
15--27$\times$ the noise floor for $k \ge 8$, while that of $\hat g_{\text{RF}}$ stays
at 0.9--1.7; with similar variances, $\hat g_{\text{RF}}$'s MSE is 2.5$\times$ lower at
$k = 16$ and $31$. At the step-25 checkpoint of a GRPO run, updated with GRPO's own
loss and optimizer state, $\hat g_{\text{IS}}$ reaches 9--10$\times$ the floor at
$k = 16, 31$, against 1.4 and 3.9 for $\hat g_{\text{RF}}$, and $\hat g_{\text{RF}}$'s
MSE is 1.3$\times$ lower; at $k = 4$--$8$, however, $\hat g_{\text{IS}}$ has the lower
MSE, because its variance is lower. The ratio-free estimator thus wins once drift
is large (here $k \ge 16$; MSE up to 2.5$\times$ lower), not at every drift level ---
the regime that the small-\texttt{mbs} experiments of \cref{sec:experiments} probe.

\paragraph{Restoring rollout drift.} In practice the rollouts come from
$\pi^*_{\text{old}} \neq \pi_{\text{old}}$ and we deploy $\pi^*_\theta$; repeating
the analysis with $\pi \to \pi^*$, the drift that governs \opft{}'s bias becomes the
\emph{deployed drift} $\pi^*_\theta - \pi^*_{\text{old}}$. Computing it would require
reloading the rollout engine after each update, so we estimate it: the decoupled
anchor $\Delta(y) = \pi_\theta - \pi_{\text{old}}$ is the faithful, cheaper estimate,
whereas the bypass anchor $\Delta^*(y) = \pi_\theta - \pi^*_{\text{old}}$ is
cheapest but mixes the two engines and is noisier --- why bypass is the less stable
trust region. GRPO's bypass-mode ratio
$\pi_\theta/\pi^*_{\text{old}}$ carries the same cross-engine mix but, being
multiplicative, turns it into exploding variance --- the worst collapse
(\cref{sec:drift_bypass}).

\subsection{Three mechanisms for stability and quality}
\label{sec:method_mechanisms}

The surrogate $J_{\text{RF}}$ alone does not guarantee stability --- with the wrong
choice of $A(y)$ the loop still collapses (the naive RS-SFT case is
the simplest example). Three mechanisms, each motivated by the
analysis above, make $J_{\text{RF}}$ both non-degenerate and faithful to $J$.

\paragraph{M1: z-score advantage reweighting.}
For prompt $x_i$ and rollout $y_{i,j}$, set $A_{i,j} = \frac{R(x_i, y_{i,j}) - \mu_i}{\sigma_i}$, where
$\mu_i, \sigma_i$ are the mean and (population) standard deviation of rewards
within the prompt's group of $N$ rollouts; prompts with $\sigma_i = 0$
are skipped. The per-prompt total of the positive $A_{i,j}$ is bounded by $N/2$
(as is the absolute total of the negatives).
The mass peaks at uncertain prompts ($k = N/2$ correct out of $N$).
M1 thus \emph{balances the
contribution from each prompt}: no easy (or hard) prompt can dominate the
update, and genuinely uncertain prompts --- where the learning signal
is richest --- contribute the most. This is exactly GRPO's
group-relative advantage estimator.
In \opft{} we additionally scale $A_{i,j}$ by $2/N$, so that the bound on the
per-prompt total positive (and total negative) advantage is $1$, decoupling it
from $N$. This is a constant rescaling of the gradient and does not change
training under a scale-invariant optimizer such as AdamW; we verify this
empirically in \cref{app:scale2n}.

\paragraph{M2: negative samples (mean-zero advantage).}
Because $A_{i,j}$ is centered per group, $\sum_j A_{i,j} = 0$ and positive-advantage samples are balanced by
negative-advantage ones, whose MLE term \emph{maximizes} NLL, pushing
the policy away from below-mean outputs. This is required by
the surrogate itself: with $A^+ = \max(A, 0)$ the objective reduces to
$J_{\text{RF}}^+ = \mathbb{E}_{\pi^*_{\text{old}}}[A^+ \log \pi_\theta]$, whose
optimum on a batch re-weights the observed positives, $\pi_\theta(y_j) \propto
A^+_j$, with no mass elsewhere; refitting round after round on samples of the
previous fit can only shrink this support, drifting toward a point mass ---
the formal counterpart of the iterative RS-SFT collapse. Mean-zero advantages (M2)
remove this degenerate fixed point. They do not make the loss bounded: a
negative-advantage term $-|A|\log\pi_\theta$ grows without limit as
$\pi_\theta \to 0$. What bounds each \emph{update} is (i) the probability floor --- a
token with $\pi_{\text{old}} \le \delta$ cannot fall by more than $\delta$; (ii) M3 on
already-likely tokens; and (iii) the gradient of $\log\pi_\theta$ with respect to the
logits, $e_{y_t} - \pi_\theta(\cdot)$, whose norm is at most $\sqrt{2}$ however small
$\pi_\theta(y_t)$ becomes.

\paragraph{M3: absolute per-token trust region.}
Because $J_{\text{RF}}$'s gradient is a sum of per-token score-function
gradients (\cref{sec:method_objective}), we can impose the trust region
independently on each token. We mask the gradient on token $t$ when the per-token
policy drift $\Delta(y_t) = \pi_\theta(y_t) - \pi_{\text{old}}(y_t)$ ---
the token-level counterpart of the bias deviation in
\eqref{eq:rfpo_bias} --- exceeds $\delta$ in the advantage-improving
direction:
\begin{equation}
\rho_{i,j,t} = 0 \ \text{ if } \operatorname{sign}(A_{i,j})\,\Delta(y_t) > \delta, \text{ and } 1 \text{ otherwise.}
\label{eq:clipmask}
\end{equation}
Positives clip the upper bound and negatives the lower; a negative token with
$\pi_{\text{old}}(y_t) \le \delta$ cannot fall by more than $\delta$ anyway, so the
mask is needed only on already-likely tokens --- elsewhere the probability floor
keeps the drop within $\delta$.
Capping $\Delta$ locally controls the policy-drift bias of \eqref{eq:rfpo_bias}.
M3 is a \emph{local} regularizer, not a sequence-level bound: after 31 inner updates
per-token ratios stay within about 1\% of 1, yet the mean absolute summed log-ratio per
sequence reaches 8.6--11.7 at our step-25 checkpoints (99th percentile up to 81;
\cref{app:gradprobe}). The anchor $\pi_{\text{old}}$ is the \emph{decoupled} (default)
one, and it is cheap ---
the actor is already loaded, so we only recompute the snapshot.
Absolute clipping is scale-independent on the per-token probability (rare tokens get the
full upward $\delta$ budget, unlike ratio clipping whose budget shrinks
linearly in $p^{\text{old}}$; \cref{sec:abs_vs_ratio} compares both
empirically and analytically).

GRPO uses the same z-score advantage (M1), trains on all samples
including $A < 0$ ones (M2), and bounds the importance ratio via
ratio clipping (M3); its KL penalty is auxiliary, and removing it does not prevent
GRPO's collapse under drift (\cref{sec:drift_mbs}). The two differ not in
\emph{which} mechanisms they use but in \emph{how} they realize them.
In the strict on-policy setup (\texttt{mbs} $=$ \texttt{bs}),
we have $r_t \equiv 1$, GRPO's clipped surrogate reduces to $A$ on every
token, and \opft{}'s clip mask is identically inactive. With the same
token-mean loss aggregation (which both our implementations use) and no KL term,
the per-token gradients are then identical up to a constant factor (\cref{app:scale2n}); our GRPO runs keep verl's $\beta=0.001$, the only other
difference at $K=1$. %
The two strategies agree in this trivial regime and diverge as
$\pi_\theta$ drifts off $\pi_{\text{old}}$ in the practical regime
(\cref{sec:method_biasvar}).

\subsection{The \opft{} algorithm}
\label{sec:method_algorithm}

\opft{} plugs into \cref{alg:opft_loop} by specifying $\textsc{Update}$.
\Cref{alg:opft} defines it: M1--M3 compose into a single mini-batch
update routine. The advantages in step~1 are computed once per training
batch (over the full set of $N$ rollouts per prompt) and reused across
mini-batches; the clip mask in step~2 is recomputed per mini-batch
since $\pi_\theta$ has drifted off $\pi_{\text{old}}$ since the
snapshot.

\begin{algorithm}[t]
\caption{$\textsc{Update}_{\opft{}}(\pi_\theta, \pi_{\text{old}}, m)$
--- the \opft{} update routine for \cref{alg:opft_loop}}
\label{alg:opft}
\begin{algorithmic}[1]
\REQUIRE Policy $\pi_\theta$, snapshot $\pi_{\text{old}}$, mini-batch
         $m$, clip threshold $\delta$, learning rate $\eta$
\STATE \textbf{Advantages (M1, M2).} For each prompt $i$ in $m$: skip
       if $\sigma_i = 0$; else
       $A_{i,j} \gets  \frac{2}{N}\frac{R(x_i, y_{i,j}) - \mu_i}{\sigma_i}$.
\STATE \textbf{Clip mask (M3).} For each token, set $\rho_{i,j,t} = 0$
       if $\operatorname{sign}(A_{i,j})\,\Delta(y_{i,j,t}) > \delta$, else $\rho_{i,j,t} = 1$; bypass
       mode replaces $\Delta$ with $\Delta^*$.
\STATE \textbf{Weighted-MLE loss.} $\mathcal{L} \gets -\sum_{i,j,t} \rho_{i,j,t}\,
       A_{i,j} \log \pi_\theta(y_{i,j,t})$, a token mean over $m$.
\STATE \textbf{Gradient step.} $\theta \gets \theta - \eta \nabla_\theta
       \mathcal{L}$.
\STATE \textbf{return} $\pi_\theta$.
\end{algorithmic}
\end{algorithm}

Common PPO/GRPO components absent in \opft{}: no reference model, no importance ratio
$\pi_\theta / \pi_{\text{old}}$, no KL term, no value function (PPO only).
$\pi_{\text{old}}$ enters only through the clip mask --- a binary
$\{0,1\}$ token decision --- so no gradient flows through it.
\cref{alg:opft_loop} with this $\textsc{Update}$ defines the full
\opft{} training procedure.

\section{Experiments}
\label{sec:experiments}

We train Qwen3-4B (instruct) \citep{qwen2025qwen3} on POLARIS-53K for math, and on DeepCoder verified 16K (PrimeIntellect's SYNTHETIC-1) for coding, \texttt{bs}=256 prompts, $N = 16$ rollouts at temperature 1.0, max
response length 4096/8192 for math/coding, lr $= 10^{-6}$, binary reward (1 if the boxed answer
matches ground truth for math, or the code passes the unit tests for coding); see \cref{app:details} for the full training configuration and the verl patch. We sweep \texttt{mbs} from 256 (\texttt{mbs}$=$\texttt{bs}, $K=1$, strict
on-policy) down to 8 ($K=32$, heavy policy drift).
During training we validate the math-trained model on a fixed, uniformly random
100-problem subset of MATH-500 \citep{hendrycks2021measuring} (``MATH-500 (100),
validation subset during training'') and the coding-trained model on LiveCodeBench-V6
(LCBv6). The subset is used only for training curves, time-to-target and the appendix
ablations, whose original checkpoints are not available for full evaluation; all other
math results use final (step-208) checkpoints on all 500 problems and AIME 2024/2025. Validation uses greedy decoding; for AIME we also report
avg@16 (16 samples at temperature 0.6, top-p 0.95), the AIME metric of
\cref{tab:full_eval} (greedy AIME per run in \cref{tab:full_eval_runs}). ``Three runs'' denotes three independent runs of the same
configuration, reported as mean $\pm$ standard deviation.

\subsection{Stability test 1: policy drift (\texttt{mbs} sweep)}
\label{sec:drift_mbs}

At strict on-policy (\texttt{mbs}$=$\texttt{bs}$=256$, $K=1$) the two updates are equivalent by construction (\cref{sec:method_mechanisms}), and GRPO serves as the
strict-on-policy reference (0.879/0.67 on MATH-500/LCBv6). The question
is whether \opft{} retains this as drift grows: smaller \texttt{mbs} $=$ larger $K$
$=$ more policy drift, the regime in which the importance-weighted estimator's error
grows (\cref{sec:method_biasvar}).

\begin{figure}[!t]
\centering
\includegraphics[width=\columnwidth]{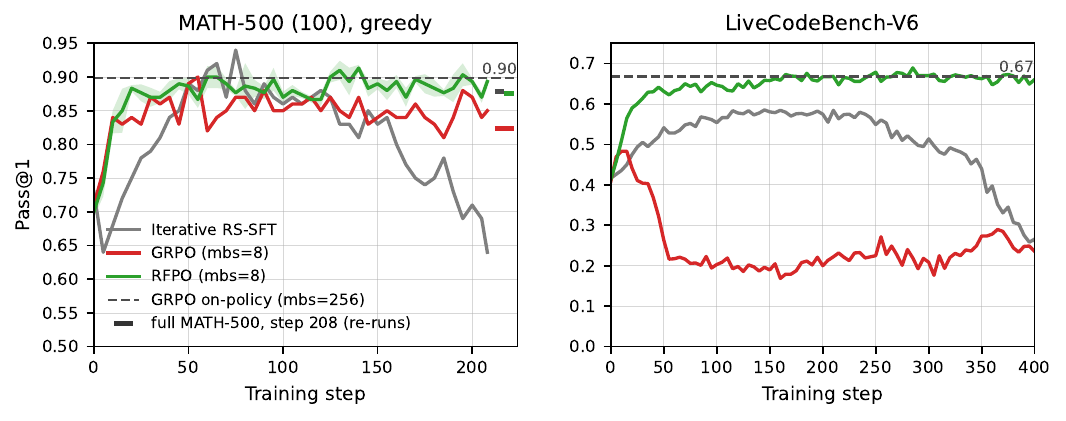}
\caption{Within-batch drift at \texttt{mbs}=8 ($K=32$). Validation
MATH-500 (100) and LCBv6 (greedy) over training. \opft{} (green) holds the
strict-on-policy target (dashed line, GRPO \texttt{mbs}=256), while GRPO (red)
degrades on math and \emph{collapses} on coding, and the positive-only iterative
RS-SFT baseline (gray) peaks then declines. The \opft{} math curve shows the band over
three runs. Math: MATH-500 (100), validation subset during training (original runs;
about $\pm$3 points per point); right-edge ticks: full MATH-500 of the re-run final
checkpoints (\cref{tab:drift_mbs}).}
\label{fig:drift_mbs8}
\end{figure}

\begin{table}[!t]
\caption{Policy drift (the \texttt{mbs} sweep) and rollout drift (bypass). \emph{Default}
columns sweep \texttt{mbs} (down the table $=$ more policy drift); \emph{bypass} columns
add rollout drift at matched \texttt{mbs}. Math: re-run checkpoints at step 208 on the
full MATH-500 (greedy); $\pm$ is the standard deviation over three runs, and
cells without it are single runs. LCBv6: original single runs, late-mean over steps
240--270 ($\sim$4.5 epochs). Per-run values and AIME are in \cref{app:fullresults}
(\cref{tab:retrain_runs}). GRPO's \texttt{mbs}=8 LCBv6 bypass cell is mid-collapse in
this window (\cref{sec:drift_bypass}).}
\label{tab:drift_mbs}
\centering
\small
\setlength{\tabcolsep}{4pt}
\begin{tabular}{l r r cc | cc}
\toprule
 & & & \multicolumn{2}{c}{default} & \multicolumn{2}{c}{bypass} \\
\cmidrule(lr){4-5} \cmidrule(lr){6-7}
Method & \texttt{mbs} & $K$ & MATH-500 & LCBv6 & MATH-500 & LCBv6 \\
\midrule
GRPO    & 256 & 1  & \textbf{0.879} {\scriptsize$\pm$0.005} & 0.667 & --- & --- \\
GRPO    & 32  & 8  & 0.861 {\scriptsize$\pm$0.015} & 0.384 & 0.718 & 0.041 \\
GRPO    & 8   & 32 & 0.824                         & 0.235 & 0.786 & 0.294 \\
\midrule
\opft{} & 32  & 8  & 0.873 {\scriptsize$\pm$0.012} & \textbf{0.672} & \textbf{0.866} {\scriptsize$\pm$0.005} & \textbf{0.669} \\
\opft{} & 8   & 32 & 0.875 {\scriptsize$\pm$0.006} & 0.669 & 0.794 & 0.568 \\
\bottomrule
\end{tabular}
\end{table}

The pattern in \cref{tab:drift_mbs} matches \cref{sec:method_biasvar}'s
prediction. At low drift the methods match, and across three math runs this
is robust rather than a one-off: \opft{} reaches $0.875\pm0.006$ (\texttt{mbs}=8)
and $0.873\pm0.012$ (\texttt{mbs}=32) against strict-on-policy GRPO's
$0.879\pm0.005$, with overlapping intervals (full MATH-500). At
matched mid-drift \texttt{mbs}=32, GRPO falls $\sim$2 points below strict on-policy in math (to 0.861; \opft{}'s 0.873 is within noise of both) and $\sim$28 points in coding, while \opft{}
holds at both \texttt{mbs}; this cleanly isolates the IS-vs-surrogate trade-off
because both methods see the same policy drift. At heavy drift
\texttt{mbs}=8, GRPO degrades a further $\sim$4 points in math (to 0.824, the clear math separation) and $\sim$15 more in coding (to 0.24), while \opft{} stays at the on-policy level (\cref{fig:drift_mbs8}). 
\cref{fig:lcb-sweep} directly links GRPO's performance drop to its rising policy
drift $\pi_{\theta}-\pi_{\text{old}}$ as \texttt{mbs} shrinks. The KL term is not
what drives it: an otherwise identical coding run at \texttt{mbs}=8 with $\beta=0$
collapses to the same level ($0.214$ vs.\ $0.206$, against \opft{}'s $0.644$).
Two further runs per method at \texttt{mbs}=8 reproduce this: GRPO with KL collapses
to 0.12 and 0.23 on LCBv6 by steps 90--110, while \opft{} reaches 0.67 and 0.66
(median over steps 240--270; base model 0.41; \cref{app:fullresults}).

\paragraph{Full MATH-500 and AIME.} \Cref{tab:full_eval} evaluates the final
checkpoint of every math arm on all 500 MATH-500 problems and on AIME 2024/2025. Over
three independent runs at \texttt{mbs}=8 (consistent with the \cref{tab:drift_mbs} re-runs), \opft{} reaches $0.869 \pm 0.011$ on MATH-500 and GRPO
$0.823 \pm 0.012$, a gap of 4.6 points, while strict on-policy GRPO reaches 0.874.
\opft{} thus matches strict on-policy accuracy rather than exceeding it (AIME 2025
avg@16: 0.344 vs.\ 0.340; GRPO at \texttt{mbs}=8: 0.232); drift tolerance buys reaching
it with 32 optimizer steps per batch, and so sooner (\cref{sec:summary}). The clip threshold matters little: $\delta$ = 0.05 / 0.1 / 0.2
gives 0.874 / 0.869 / 0.858.

\begin{table}[t]
\caption{Final math checkpoints (step 208) on all 500 MATH-500 problems (greedy) and on
AIME 2024/2025 (avg@16; greedy AIME and the $\delta$ sweep are in
\cref{tab:full_eval_runs}). \texttt{mbs}=8 ($K=32$) unless noted; default anchor.
$\pm$: standard deviation over three runs; other rows are single runs. GSPO and DAPO
(without dynamic sampling) are run at the same settings; per-run values are in
\cref{app:fullresults}.}
\label{tab:full_eval}
\centering
\small
\setlength{\tabcolsep}{4pt}
\begin{tabular}{l c c c c}
\toprule
Method & \texttt{mbs} & MATH-500 & AIME-24 & AIME-25 \\
\midrule
Base model (Qwen3-4B) & --- & 0.678 & 0.117 & 0.094 \\
GRPO (strict on-policy, $\beta$=0.001) & 256 & 0.874 & 0.417 & 0.340 \\
\midrule
GRPO ($\beta$=0.001, three runs) & 8 & 0.823 {\scriptsize$\pm$0.012} & 0.302 & 0.232 \\
GRPO ($\beta$=0) & 8 & 0.818 & 0.302 & 0.244 \\
GSPO & 8 & 0.702 & 0.102 & 0.085 \\
DAPO (no dynamic sampling) & 8 & 0.858 & 0.323 & 0.298 \\
\midrule
\opft{}, $\delta$=0.1 (three runs) & 8 & 0.869 {\scriptsize$\pm$0.011} & 0.396 & 0.344 \\
\bottomrule
\end{tabular}

\vspace{8pt}
\caption{Objective $\times$ trust region at \texttt{mbs}=8, no KL. Math: full
MATH-500, step 208. Coding: LCBv6, median over steps 240--270 (base 0.41); medians
exclude isolated single-point evaluation failures of the ratio-free runs (means in
\cref{app:fullresults}). The GRPO coding runs collapsed by steps 90--110 and were
stopped. $^*$With its KL term ($\beta = 0.001$); KL-free GRPO also collapses (0.214).}
\label{tab:twobytwo}
\small
\begin{tabular}{l cc cc}
\toprule
 & \multicolumn{2}{c}{ratio clip ($\epsilon = 0.2$)} & \multicolumn{2}{c}{absolute clip ($\delta = 0.1$)} \\
\cmidrule(lr){2-3} \cmidrule(lr){4-5}
Objective & math & coding & math & coding \\
\midrule
importance ratio & 0.818 & collapses (0.12--0.23)$^*$ & 0.860 & collapses (0.00) \\
ratio-free & 0.858 & 0.65 & 0.869 {\scriptsize$\pm$0.011} & 0.66--0.67 (\opft{}) \\
\bottomrule
\end{tabular}
\end{table}

\subsection{Stability test 2: rollout drift (bypass mode)}
\label{sec:drift_bypass}

The default recipe recomputes $\pi_{\text{old}}$ with the training engine, so the
actor and its clip anchor share numerics --- the faithful setting of
\cref{sec:method_biasvar}. Verl's \emph{bypass} mode instead reuses vLLM's
rollout log-probabilities as $\pi_{\text{old}}$: faster, since it skips a forward
pass over every rollout, but no longer the correct trust-region anchor --- it
substitutes $\pi^*_{\text{old}}$ for $\pi_{\text{old}}$, mixing the two engines.
The \emph{bypass} columns of \cref{tab:drift_mbs} report both methods at matched
\texttt{mbs}.

Both pay for the mixed engine, but at \texttt{mbs}=32 very differently: \opft{} is
nearly unchanged ($0.873\to0.866$ MATH-500, $0.672\to0.669$ LCBv6) while GRPO
collapses, by $\sim$14 points on MATH-500 ($0.861\to0.718$) and to
near zero on LCBv6 (\cref{fig:bypass_curves}; both GRPO coding runs are below
$0.05$ at step 400, \texttt{mbs}=8 collapsing later (crossing $0.1$ near step 360 vs.\ 215),
so their order in \cref{tab:drift_mbs} reflects timing, not outcome). At \texttt{mbs}=8,
where bypass adds to 32 steps per batch, the combined drift hurts both on math: \opft{}
falls from 0.875 to 0.794 and GRPO from 0.824 to 0.786 (one run each), about the same
accuracy; the separation remains on coding (0.568 vs.\ 0.294). In bypass mode GRPO's ratio $r_t = \pi_\theta/\pi^*_{\text{old}}$ divides by
the \emph{actual} rollout distribution, more faithful than default mode. But GRPO's
multiplicative ratio turns the cross-engine mismatch into variance, whereas
\opft{}'s ratio-free gradient confines it to the binary clip mask. Default is
thus the mode to prefer, but \opft{} can also take the bypass speedup
when throughput matters --- at a cost that grows as \texttt{mbs} shrinks --- an
option GRPO does not have.

\subsection{Removing the ratio or changing the trust region?}
\label{sec:twobytwo}

\opft{} differs from GRPO in two ways: the objective (GRPO's importance-weighted
gradient $r_t A \nabla\log\pi_\theta$ vs.\ \opft{}'s ratio-free $A\nabla\log\pi_\theta$)
and the trust region (PPO's clip of $r_t$ to $1 \pm 0.2$ vs.\ \opft{}'s mask on the
absolute change in token probability, $\delta = 0.1$). \Cref{tab:twobytwo} crosses the
two at \texttt{mbs}=8 without KL. On math, either change closes most of the gap, and
only GRPO's combination falls behind. On coding, where drift is heavier, the objective
decides: both cells with the ratio collapse whichever clip they use, and both
ratio-free cells stay stable. Under heavy drift, the ratio is what breaks. The two
added baselines behave like the importance-weighted cells: DAPO (without dynamic
sampling, which our trainer does not support) reaches 0.858 on math and collapses on
coding (0.09 by step 90); GSPO reaches 0.702 and 0.27.

\subsection{Mechanism ablations}
\label{sec:ablation}

\Cref{tab:ablation_loo} (\cref{app:figures}) tests each mechanism from both
directions: building \opft{} up from positive-only iterative RS-SFT one mechanism at a
time, and removing one mechanism from the full recipe (late-means on MATH-500 (100), validation subset during training).
\textbf{M1 (z-score) smooths optimization.} Adding it to RS-SFT lowers grad-norm CoV
($0.34\to0.29$) and lifts accuracy ($0.71\to0.78$); removing it from the full recipe
raises CoV $2.5\times$ ($0.18\to0.46$) but costs only $\sim$2 points, and the $+$M1 run
still collapses from its peak (\cref{fig:cumulative_addone}): M1 is mainly a quality
mechanism, the counterpart of GRPO's group-relative advantage. \textbf{M2 (negatives)
and M3 (clipping) together provide stability.} M2 without M3 \emph{worsens} collapse
($0.61$, CoV $4.97$): negatives without a trust region oscillate with unbounded
magnitude, and M3 bounds those updates ($0.91$). Removing M2 from the full recipe
reproduces the RS-SFT collapse ($0.59$), silent in grad-norm (CoV $0.33$); removing M3
reopens the exploding-variance collapse. \textbf{Failed-stabilizer audit.} None of six standard regularizers (entropy, KL to a
reference, EMA rollouts, repetition penalties, ratio or absolute clipping) added to
positive-only RS-SFT prevents collapse (\cref{fig:failed_stabilizers}).

\subsection{Efficiency}
\label{sec:summary}

\opft{} skips the reference forward, the importance-ratio computation, and the KL
loss. We time every arm on the same single 8$\times$H100 node at \texttt{mbs}=8
(median seconds per step; one step is one outer batch of 256 prompts $\times$ 16
rollouts, including generation and all inner optimizer steps). On math: \opft{} 367,
ratio-free $+$ ratio clip 394, ratio $+$ absolute clip 421, KL-free GRPO 514, GRPO
with KL 586 (FSDP2), and strict on-policy GRPO (\texttt{mbs}=256) 514. At matched
\texttt{mbs}=8, \opft{} is thus 1.4$\times$ faster per step than KL-free GRPO and
1.6$\times$ faster than GRPO with KL. On coding: \opft{} 756--773,
GRPO with KL 1585--1691, DAPO 1427. The GRPO-with-KL arms run FSDP2 (needed for the
KL term at small \texttt{mbs} on our stack) and the others FSDP1; since wall-clock
includes generation, it also reflects response length, which we do not separate from
update time. Drift tolerance also cuts the number of steps: \opft{} at
\texttt{mbs}=8 first reaches 0.85 on MATH-500 (100) (validation subset during training) within 10--20 steps
($\sim$1.9\,h), against 30 steps ($\sim$5.7\,h) for strict on-policy GRPO, and
0.60 on LCBv6 at step 25 (step 65 at \texttt{mbs}=32), against 135 for strict
on-policy GRPO; GRPO at \texttt{mbs} 32 or 8 never reaches 0.60, with or without KL.

\section{Conclusion and Discussion}
\label{sec:discussion}
\label{sec:conclusion}

We framed on-policy fine-tuning through a bias-variance lens: GRPO
importance-corrects and pays an error that grows with drift, while \opft{}
retargets to the ratio-free surrogate $J_{\text{RF}}$ and pays a
trust-region-controlled bias. A direct measurement shows the ratio-free gradient has
the lower MSE once drift is large; GRPO degrades or collapses while \opft{} stays at the
on-policy level under policy drift and under rollout drift at 8 steps per batch, with its three mechanisms playing distinct roles. RL-style training without a reference model, KL penalty, or importance ratio is
thus feasible at the small \texttt{mbs} that enables many optimizer steps per batch.

\textbf{Limitations and scope.}
Our study covers Qwen3-4B on two verifiable-reward domains (math, POLARIS-53K;
coding, DeepCoder-16K); the second backbone, Phi-4-mini-reasoning
(\cref{app:phi4mini}), covers coding only, so math rests on a single backbone; other scales and task
families are untested. GRPO at \texttt{mbs}=8, the \texttt{mbs}=8 bypass arms and most coding arms are single runs; DAPO
runs without dynamic sampling; we do not compare with MinPRO or LGPO; the gradient
probe uses two early (step-25) checkpoints and one reference draw per inner step.

\textbf{Next steps.} A sequence- or prefix-level trust region on the ratio-free
surrogate (\cref{app:future_work}), and long-horizon agentic training, where both
drifts compound across turns.

\section*{Reproducibility Statement}
All experiments use the public Qwen3-4B model and the POLARIS-53K (math) and
DeepCoder-16K (coding) datasets (\cref{sec:experiments}). \opft{}'s full update
rule is given in \cref{alg:opft} (\cref{sec:method_algorithm}), and its three
mechanisms M1--M3 --- z-scored advantages, negative samples, and an
absolute per-token trust region ($\delta=0.1$) --- are defined in
\cref{sec:method_mechanisms}. Complete training hyperparameters for both \opft{}
and GRPO, together with the verl offload-symmetry patch required to reproduce the
small-\texttt{mbs} GRPO runs, are in \cref{app:details} (\cref{tab:hparams}); the
bias-variance derivation is in \cref{app:theory_proofs} and the gradient-estimator
probe protocol in \cref{app:gradprobe}. Evaluation uses MATH-500 (all 500 problems
for final checkpoints; a fixed 100-problem validation subset only for training curves, time-to-target and the appendix ablations), LiveCodeBench-V6,
and AIME 2024/2025 under greedy decoding, with avg@16 also reported for AIME. We will release our training code and configurations.

\section*{AI Use Statement}
In this work, we used generative AI tools for two purposes, both of which fall
under the \emph{recommended} disclosure category of the ICLR 2027 AI policy.
First, we used LLM-based assistants to aid and polish writing: editing prose for
clarity and concision, on text whose content, claims, and organization were
authored by us. Second, we used LLM-based tools for retrieval and discovery, in
particular for searching for related work and surfacing candidate references;
every reference so surfaced was located in its original form, read by the
authors, and verified before citation.

We did not use generative AI tools for tasks in the policy's \emph{required}
disclosure category. The research question and the \opftloop{} framing, the
\opft{} objective and its three mechanisms, the bias-variance analysis and its
derivation (\cref{app:theory_proofs}), the experimental design, the
implementation of \opft{} and the verl offload-symmetry patch, and the
interpretation of all reported results are the authors' own work.

We have reviewed all AI-assisted work: AI-edited prose was checked line by line
against the authors' intended meaning, and every AI-suggested reference was
independently retrieved and read before being cited. We take responsibility for
the final content of this work, including text, claims, or artifacts produced
with the aid of generative AI.

\bibliography{references}
\bibliographystyle{iclr2027_conference}

\appendix

\section{Experiment details}
\label{app:details}

\Cref{tab:hparams} lists the shared training configuration; the two methods are
identical except the rows noted. Rollouts are generated by vLLM and training runs
under FSDP, and validation uses greedy decoding, late-meaned for the training curves
and the LCBv6 cells of \cref{tab:drift_mbs} over the windows given there; the math cells
of \cref{tab:drift_mbs} are step-208 re-run checkpoints (\cref{app:fullresults}). The final evaluations (\cref{tab:full_eval,app:fullresults}) use
the same validation code on all 500 MATH-500 problems (greedy) and on AIME 2024/2025
(greedy and avg@16).

\begin{table}[t]
\caption{Training hyperparameters. Spanned cells are shared by both methods;
only the three rows below the second rule differ. The Phi-4-mini-reasoning runs
of \cref{app:phi4mini} are coding-only and train for 406 steps ($\approx$7
epochs); every other entry is as listed.}
\label{tab:hparams}
\centering
\small
\setlength{\tabcolsep}{4pt}
\begin{tabular}{@{} p{0.22\columnwidth} p{0.36\columnwidth}
                    p{0.36\columnwidth} @{}}
\toprule
                 & \opft{}                 & GRPO \\
\midrule
Base model       & \multicolumn{2}{@{}p{0.74\columnwidth}@{}}{Qwen3-4B (instruct); Phi-4-mini-reasoning (\cref{app:phi4mini})} \\
Training data    & \multicolumn{2}{@{}l@{}}{POLARIS-53K (math)/DeepCoder-16K (coding)} \\
Steps            & \multicolumn{2}{@{}l@{}}{208 (math, 1 epoch); coding $\sim$4.5 epochs} \\
Train batch      & \multicolumn{2}{@{}l@{}}{256 prompts} \\
Rollouts $N$     & \multicolumn{2}{@{}l@{}}{16 @ T=1.0} \\
Max resp.\ len.  & \multicolumn{2}{@{}l@{}}{4096 (math)/8192 (coding)} \\
LR               & \multicolumn{2}{@{}l@{}}{$10^{-6}$} \\
Negatives       & \multicolumn{2}{@{}l@{}}{weight $1.0$, symmetric (no downweighting)} \\
Loss aggregation & \multicolumn{2}{@{}l@{}}{token-mean over the mini-batch (verl \texttt{loss\_agg\_mode})} \\
\midrule
Clip mode        & absolute, $\delta=0.1$  & ratio, $\epsilon=0.2$ \\
KL penalty       & ---                     & $\beta=0.001$ ($\beta=0$ control: \cref{sec:drift_mbs}) \\
Reference model  & none                    & frozen copy of init \\
\bottomrule
\end{tabular}
\end{table}

\paragraph{Engines and the rollout-training gap.} Rollouts are sampled by vLLM in
bf16; the actor trains under FSDP with bf16 compute and fp32 master weights. GRPO runs
with a KL term ($\beta=0.001$), in \cref{tab:drift_mbs} and \cref{tab:full_eval}, use
verl's FSDP2 backend; all other arms (all \opft{} runs, KL-free GRPO, GSPO, DAPO, ratio
$+$ absolute clip) use FSDP1. Both shard the same bf16 computation; \cref{sec:summary}
notes the step-time effect. The default anchor
$\pi_{\text{old}}$ is recomputed by the training engine; bypass mode uses vLLM's
log-probabilities and so mixes the two engines. Both engines run in bf16 but with
different kernels and reduction order, so the same weights give slightly different
probabilities. We measured the gap on the same tokens (14--17M response tokens of the
probe batches of \cref{app:gradprobe}, vLLM v0.22.1 at tensor-parallel size 1 against
the FSDP actor in bf16): mean $|\Delta\log p| = 0.014$ (99th percentile 0.19) and mean
$|\Delta p| = 0.0057$ (99th percentile 0.06); vLLM is slightly more confident (signed
mean $\Delta\log p \approx +8\times10^{-4}$). The figures are about the same at the
initial model and at both step-25 checkpoints. The vLLM attention backend is
auto-selected, and the measurement's FSDP side uses PyTorch SDPA attention rather
than the training actor's flash-attention kernel.

\paragraph{Evaluation noise.} Greedy vLLM decoding is not exactly repeatable across
evaluation passes (8--9 of 100 problems flip between passes). Re-evaluating the same
checkpoint changed full MATH-500 by up to 2.2 points (0.786 vs.\ 0.764) and the
100-problem subset by up to about 3 points, so differences below about 2 points between
single full-set evaluations are not meaningful.

\paragraph{verl patch.} Running GRPO at small \texttt{mbs} under FSDP with
parameter offload deadlocked in the reference-model forward: verl offloads the
reference module but --- unlike the actor --- never loads it back before each
forward, so with many optimizer steps per batch one rank eventually reads a freed
shard and stalls NCCL. A $\sim$10-line fix mirroring the actor's load/offload for
the reference module (the ``patched'' verl) removes the deadlock at all
\texttt{mbs}; all GRPO runs use it, so the reported regressions reflect
algorithmic instability (the importance-weighted update under drift), not this
operational one.

\section{Bias-variance derivation}
\label{app:theory_proofs}

Throughout this appendix we adopt vanilla GRPO's implicit identity
$\pi^*_{\text{old}} = \pi_{\text{old}}$ (\cref{sec:method_biasvar}), so
the sampler coincides with $\pi_{\text{old}}$, and we idealize
$\hat g_{\text{IS}}$ as unbiased for $\nabla J$ (hence the $\|\nabla J\|^2$ term
below), ignoring the residual prefix and continuation mismatch of per-token ratios noted in
\cref{sec:method_objective}; this idealization favors GRPO. A rollout drift $\pi^*_{\text{old}} \neq
\pi_{\text{old}}$ adds bias to $\hat g_{\text{IS}}$
(\cref{sec:drift_bypass}) but leaves $\hat g_{\text{RF}}$ unchanged.

\paragraph{Variance of $\hat g_{\text{IS}}$.}
With $\hat g_{\text{IS}} = r_t \cdot A \cdot \nabla \log \pi_\theta$ and
$r_t = \pi_\theta / \pi_{\text{old}}$,
\begin{equation*}
\mathrm{Var}_{\pi_{\text{old}}}(\hat g_{\text{IS}})
= \mathbb{E}_{\pi_{\text{old}}}[r_t^2 \cdot \|A \nabla \log
  \pi_\theta\|^2] - \|\nabla J\|^2.
\end{equation*}
Ratio clipping does \emph{not} keep every ratio in $[1-\epsilon, 1+\epsilon]$. The
clipped objective $\min\bigl(r_t A,\ \mathrm{clip}(r_t, 1-\epsilon, 1+\epsilon)\,A\bigr)$
zeroes a token's gradient only once $r_t$ leaves the band in the
advantage-improving direction ($r_t > 1+\epsilon$ for $A > 0$, $r_t < 1-\epsilon$
for $A < 0$). A token with $A < 0$ whose probability \emph{rose} keeps the unclipped
term $r_t A \nabla\log\pi_\theta$ with $r_t > 1+\epsilon$. In verl the only cap on it
is the dual clip, which takes the loss constant once $r_t > c$ with $c = 3$; it is
active in our runs, firing on $9.6\times10^{-5}$ of tokens at \texttt{mbs}=8 (math,
after step 50) and never at \texttt{mbs}=256, where $r_t \equiv 1$. On a single step
the first term above is therefore bounded only by about $c^2 = 9$ times the
on-policy level $\sigma^2_{\text{on-policy}}$, not by $(1+\epsilon)^2$. Across $K$
steps within a batch, $\pi_\theta$ moves while $\pi_{\text{old}}$ is held fixed, and
the per-token amplification compounds.

\paragraph{Bias of $\hat g_{\text{RF}}$.}
The estimator $\hat g_{\text{RF}} = A \cdot \nabla \log \pi_\theta$ is unbiased for
$\nabla J_{\text{RF}}$ but biased for $\nabla J$. Writing
$\Delta^*(y) = \pi_\theta(y) - \pi^*_{\text{old}}(y)$
(which reduces to $\pi_\theta - \pi_{\text{old}}$ under this appendix's
$\pi^*_{\text{old}} = \pi_{\text{old}}$),
\begin{equation*}
\nabla J - \nabla J_{\text{RF}} = \textstyle\sum_y \Delta^*(y) \cdot A(y) \cdot
\nabla \log \pi_\theta(y),
\end{equation*}
so $\|\mathrm{Bias}(\hat g_{\text{RF}})\| \leq C \cdot \|\Delta^*\|_1 \leq 2C \cdot
\mathrm{TV}(\pi_\theta, \pi_{\text{old}})$ with $C =
\sup \|A \cdot \nabla \log \pi_\theta\|$. Absolute clipping caps the per-token
additive drift on contributing tokens to $\delta$; ratio clipping caps the
multiplicative drift to $\epsilon$. Neither cap bounds the sequence-level
$\mathrm{TV}$ by its threshold: even if every position drifted by at most $\delta$,
the telescoping bound would give only $\mathrm{TV} \le T\delta$, vacuous for long
responses, so the bound is informative only for small drift, and we read M3 as a
local regularizer of this bias rather than a global guarantee.

\paragraph{Variance of $\hat g_{\text{RF}}$.}
$\mathrm{Var}_{\pi_{\text{old}}}(\hat g_{\text{RF}}) =
\mathbb{E}_{\pi_{\text{old}}}\bigl[\|A \nabla \log \pi_\theta\|^2\bigr] -
\|\nabla J_{\text{RF}}\|^2$ carries no ratio factor, but it is not independent of
drift: $\nabla\log\pi_\theta$ is evaluated at the drifting $\theta$. What holds is
only that $\hat g_{\text{RF}}$ has no $r_t^2$ term; neither bound settles the
comparison, which is why we measure it (\cref{app:gradprobe}).

\paragraph{Clipping bias.} Both recipes also incur a \emph{clipping/masking} bias relative to their own \emph{unclipped} estimators. Practical GRPO is not strictly unbiased because ratio clipping suppresses gradient contributions once the policy moves sufficiently far in the advantage-improving direction. Relative to unclipped importance-weighted gradient:
\begin{equation*}
\text{Bias}_{\text{GRPO,clip}} = -\mathbb{E}_{\pi_{\text{old}}}[r_t A \cdot
\nabla \log \pi_\theta(y) \mathbf{1}\{A\Delta^*(y)>|A|\epsilon p_{\text{old}}\}].
\end{equation*}
For \opft{}, the bias introduced by masking relative to the unmasked surrogate is
\begin{equation*}
\text{Bias}_{\text{RFPO,clip}} = -\mathbb{E}_{\pi_{\text{old}}}[A \cdot
\nabla \log \pi_\theta(y) \mathbf{1}\{A\Delta^*(y)>|A|\delta\}].
\end{equation*}
Two differences follow. First, the trigger: For tokens with $p_{\text{old}}<\delta/\epsilon$, \opft{} has strictly larger allowable directional probability budget than GRPO's earlier clipping. With our $\delta=0.1,
\epsilon=0.2$ this is $\pi_{\text{old}}(y_t) < 0.5$, i.e.\ essentially every
token but the highest-probability ones.  Second, the weight: GRPO's discarded gradient contributions carry the multiplicative importance factor, which itself grows
with drift, whereas \opft{}'s carries none.

Note that the two clip-bias expressions above are each measured against that
method's \emph{own} unclipped estimator, so they are not directly
comparable as errors in $\nabla J$. Under the idealization above, GRPO's only bias is $\text{Bias}_{\text{GRPO,clip}}$. \opft{} trades
the $r_t^2$ variance for an \emph{additional} surrogate-gap term --- which
is precisely the term M3 controls locally. 

\paragraph{No closed-form crossover.} We do not derive when the squared bias of
$\hat g_{\text{RF}}$ falls below the error of $\hat g_{\text{IS}}$. \Cref{app:gradprobe}
measures bias, variance and MSE of both estimators directly: the ratio-free
estimator has the lower MSE once drift is large ($k \ge 16$ inner steps), while at
small drift the importance-weighted estimator is as good or better
(\cref{sec:method_biasvar}).

\section{Absolute vs.\ ratio clipping (M3)}
\label{sec:abs_vs_ratio}

Ratio clipping ($\pi_{\text{new}}/\pi_{\text{old}} \in [1-\epsilon,
1+\epsilon]$) is scale-dependent. Specifically, the allowable absolute probability movement of GRPO is $\epsilon\pi_{\text{old}}$, whereas for \opft{} it is $\delta$, and the crossover point is $\pi_{\text{old}}<\frac{\delta}{\epsilon}$. With $\delta=0.1,\ \epsilon=0.2$ this crossover is at $\pi_{\text{old}}=0.5$, so absolute clipping is the more permissive rule on essentially every token except the highest-probability ones. At $\pi_{\text{old}} = 0.001$ and
$\epsilon=0.2$, the maximum allowed $p_{\text{new}}$ is $0.0012$ ---
$500\times$ tighter than absolute clipping at $\delta=0.1$. This matters
for positives, which often act on rare tokens that should become common; for
negatives, a token with $\pi_{\text{old}} \le \delta$ cannot fall by more than
$\delta$, so the mask is needed only on already-likely tokens. Empirically, with the
ratio-free objective and all other mechanisms held fixed, absolute clipping reaches
$0.869 \pm 0.011$ on full MATH-500 against 0.858 for ratio clipping, and 0.66--0.67
against 0.65 on LCBv6 (\cref{tab:twobytwo}).

\section{Trust-region granularity and concurrent work}
\label{app:future_work}

M3 in this paper is a \emph{per-token} trust region: mask the gradient
on token $t$ when $\operatorname{sign}(A)\,(\pi_\theta(y_t) - \pi_{\text{old}}(y_t)) > \delta$.
Several concurrent works place the trust region at coarser
granularities. \citet{zheng2025gspo} (GSPO) replace token-level
importance ratios with a single sequence-level ratio.
\citet{lei2026minpro} (MinPRO) replace the cumulative prefix ratio with
the minimum token-level ratio observed in the preceding prefix.
\citet{li2025trustregion} bound the maximum token-level divergence
across a sequence and mask entire violating sequences. Per-token
clipping gives every position a separate budget and admits asymmetric
clipping (positives clip up, negatives clip down); sequence- and
prefix-level alternatives are a single budget over the whole rollout
and may behave differently when either (a) a few tokens dominate the
deviation or (b) the deviation is distributed evenly across many
tokens. All these placements serve M3's theoretical role of controlling
the bias of $J_{\text{RF}}$ relative to $J$ (\cref{sec:method_biasvar}), but they
correspond to different points on the bias-variance curve and should
be compared empirically. The per-token placement does not bound sequence-level drift:
after 31 inner updates per-token ratios stay within about 1\% of 1, while the mean
absolute summed log-ratio per sequence reaches 3.3 at the initial model, 8.6 at an
\opft{} checkpoint and 11.7 at a GRPO checkpoint, with a 99th percentile up to 81
(\cref{app:gradprobe}). M3 is therefore best read as a local regularizer, and a
sequence- or prefix-level trust region on the ratio-free surrogate, in the style of
GSPO or MinPRO, is its natural extension. An open question is whether the bias-variance
MSE prediction of \cref{sec:method_biasvar} sharpens or interacts with
\citet{li2025trustregion}'s monotonic-improvement bound when the two
are evaluated on the same recipe.

\section{Additional result figures}
\label{app:figures}

\begin{figure}[t]
  \centering
  \begin{subfigure}[b]{0.48\textwidth}
    \includegraphics[width=\textwidth]{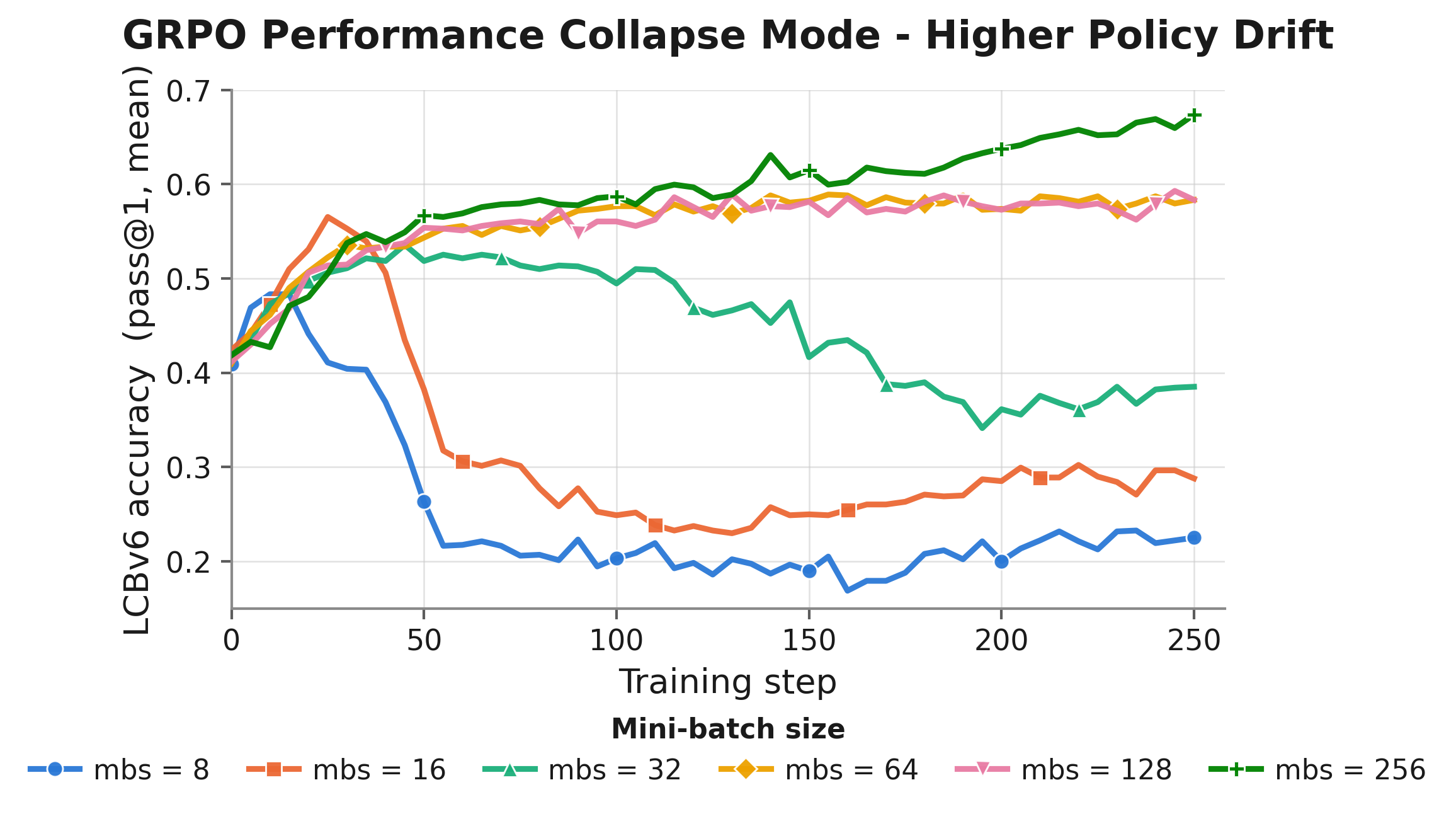}
    \caption{LCBv6 accuracy}
    \label{fig:lcb-mbs-sweep}
  \end{subfigure}
  \hfill
  \begin{subfigure}[b]{0.48\textwidth}
    \includegraphics[width=\textwidth]{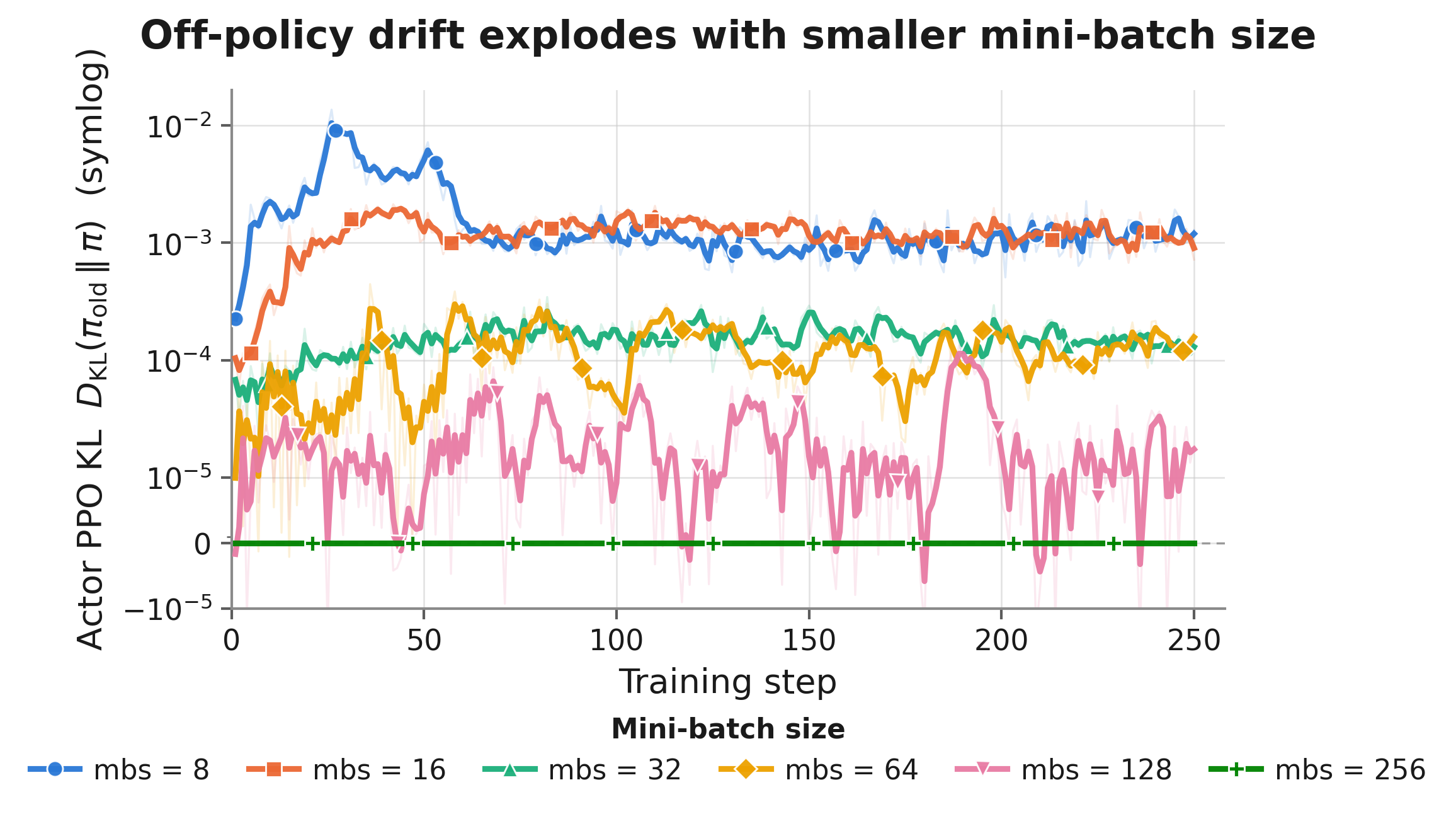}
    \caption{KL divergence between $\pi_{\theta}$ and $\pi_{\text{old}}$}
    \label{fig:lcb-mbs-sweep-kl}
  \end{subfigure}
  \caption{Coding GRPO across mini-batch sizes. As \texttt{mbs} decreases (more
  optimizer steps per batch, i.e.\ more policy drift), LCBv6 accuracy falls (a)
  in step with a rising $\pi_{\theta}$--$\pi_{\text{old}}$ KL (b) --- the
  drift that drives the importance-weighted estimator's error (\cref{sec:method_biasvar}).}
  \label{fig:lcb-sweep}
\end{figure}

\begin{figure}[h]
\centering
\includegraphics[width=0.485\columnwidth]{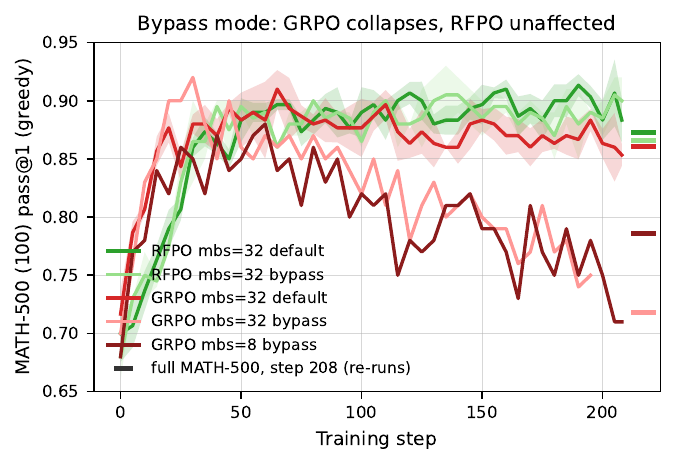}\includegraphics[width=0.485\columnwidth]{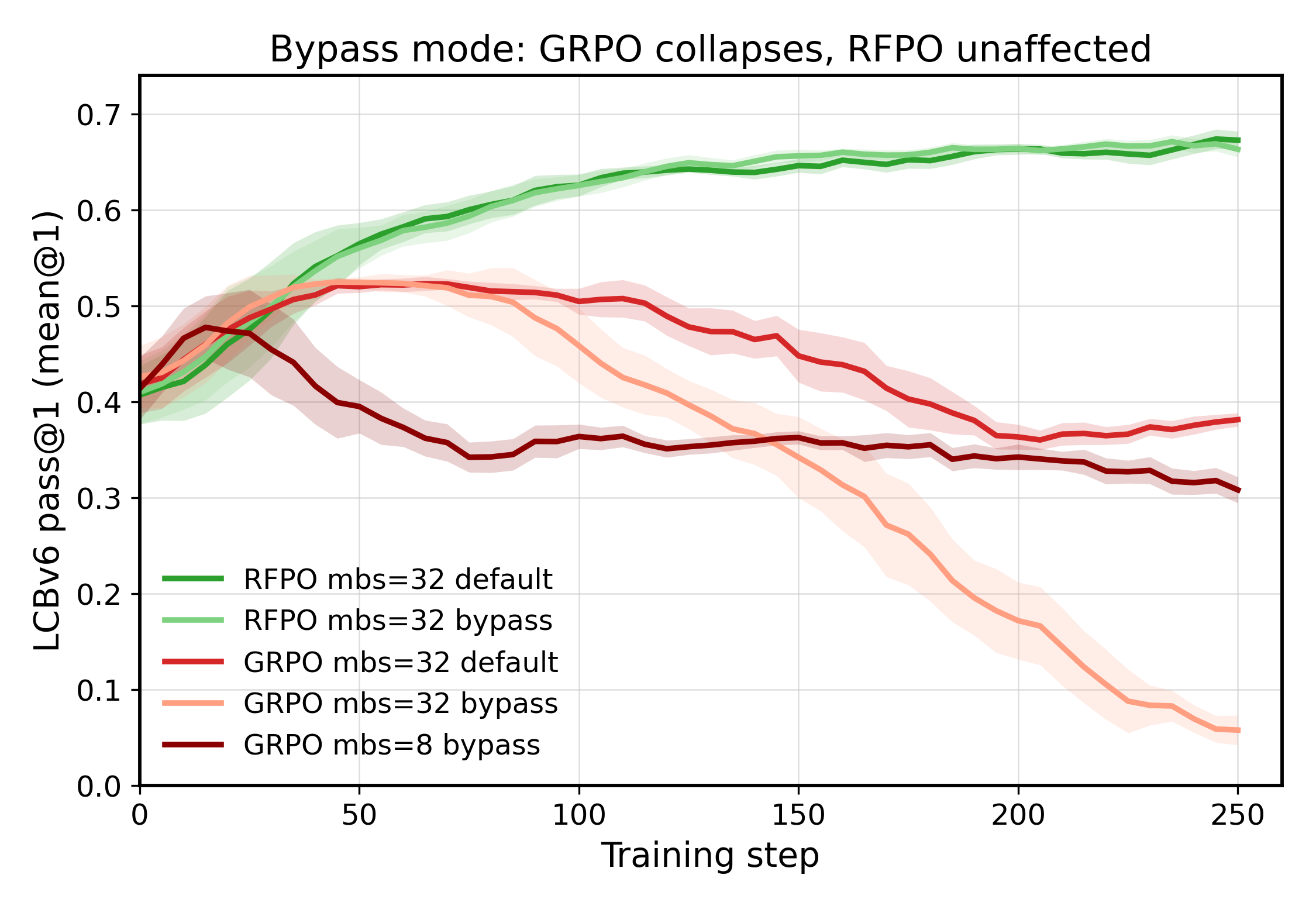}
\caption{Bypass-mode training curves on MATH-500 and LiveCodeBench~v6 (greedy pass@1; \opft{} shown at \texttt{mbs}=32). On both benchmarks \opft{} bypass stays close to \opft{} default, while both GRPO bypass runs (\texttt{mbs}=32 and \texttt{mbs}=8) peak near step~30 then decline monotonically as GRPO default sustains. On MATH-500 (100) the bypass runs fall to 0.71--0.75; on LiveCodeBench~v6 the collapse is far more severe---over steps 240--270 (the \cref{tab:drift_mbs} window) GRPO bypass averages $0.04$ at \texttt{mbs}=32 and $0.29$ at \texttt{mbs}=8, which collapses later, versus \opft{}'s stable $\approx$0.66. The collapse comes from the actor-rollout mismatch entering GRPO's \emph{multiplicative} importance ratio; \opft{}'s ratio-free gradient admits it only through a bounded clip mask.
At step 400, GRPO in bypass mode collapses further to below 0.05 for both \texttt{mbs}=32 and \texttt{mbs}=8. Math panel: MATH-500 (100), validation subset during training (original runs; about $\pm$3 points per point); ticks at the right edge: full MATH-500 of the re-run final checkpoints (\cref{tab:drift_mbs}).}
\label{fig:bypass_curves}
\end{figure}

\begin{figure}[h]
\centering
\includegraphics[width=0.5\columnwidth]{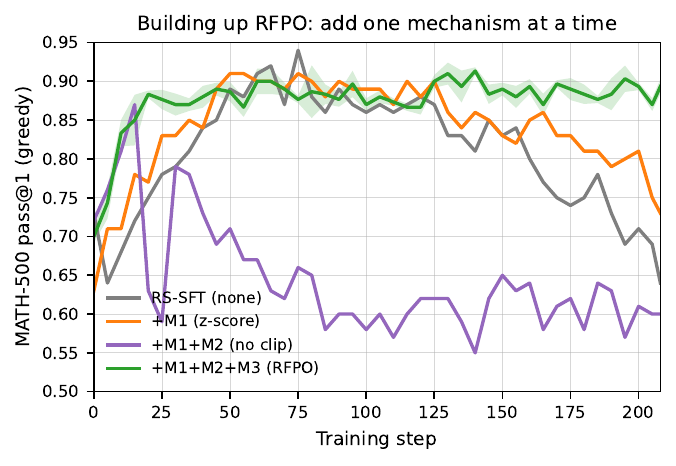}
\caption{Cumulative ablation (MATH-500 (100), validation subset during training; greedy): building \opft{} from iterative
RS-SFT by adding one mechanism at a time. M1 alone delays collapse (late-mean
$0.71\to0.78$) but does not prevent it; M2 without M3 collapses harder (to
$0.61$ --- negatives need a trust region); only the full M1$+$M2$+$M3 recipe
stays stable ($0.91$, band over three runs). Partial configs are single runs.}
\label{fig:cumulative_addone}
\end{figure}

\begin{figure}[h]
\centering
\includegraphics[width=0.48\columnwidth]{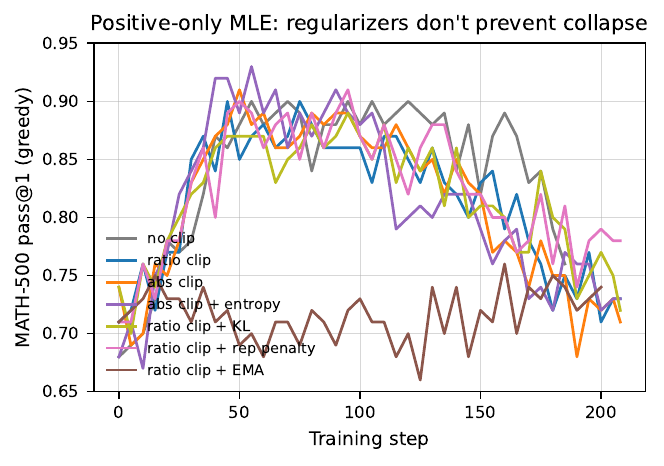}
\caption{Failed-stabilizer audit (MATH-500 (100), validation subset during training; single runs):
the positive-only iterative RS-SFT baseline plus a range of regularizers, all of
which peak then decline --- none recovers stability. Final values --- no clip
$0.76$, ratio clip ($\epsilon{=}0.2$) $0.74$, absolute clip ($\delta{=}0.1$)
$0.72$, $+$entropy $0.74$, $+$KL-to-ref $0.75$, $+$rep.\ penalty $0.78$,
$+$EMA rollout $0.73$ --- all far below the full \opft{} recipe's $\approx0.91$,
confirming that M2's corrective gradient cannot be substituted by a side
regularizer.}
\label{fig:failed_stabilizers}
\end{figure}

\begin{figure}[t]
\centering
\includegraphics[width=0.55\columnwidth]{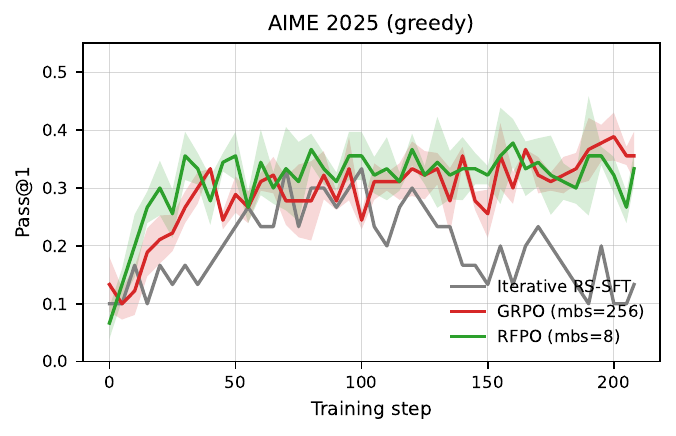}
\caption{AIME 2025 (greedy) during math training. \opft{} (\texttt{mbs}=8, band over three runs)
matches GRPO (\texttt{mbs}=256, band over three runs) while the positive-only iterative
RS-SFT baseline collapses. GRPO is shown only at strict on-policy. Final-checkpoint AIME 2024/2025 for every arm: avg@16 in \cref{tab:full_eval}, greedy in \cref{tab:full_eval_runs}.}
\label{fig:aime}
\end{figure}

This appendix collects supplementary figures for \cref{sec:experiments}.
\Cref{fig:lcb-sweep} expands the coding \texttt{mbs} sweep and \cref{fig:bypass_curves}
the bypass-mode collapse (both referenced from the main text); \cref{tab:ablation_loo} and \cref{fig:cumulative_addone,fig:failed_stabilizers}
give the mechanism ablations, and \cref{fig:aime} reports AIME 2025 (deferred
from the main text, which uses MATH-500).

\begin{table}[h]
\caption{Mechanism ablation on math training (\texttt{mbs}=8; late-mean on MATH-500 (100), validation subset during training). \emph{Top block}
builds \opft{} up from positive-only iterative RS-SFT one mechanism at a time;
\emph{bottom block} removes a single mechanism from the full recipe. M1 alone
only delays collapse ($0.71\!\to\!0.78$); adding M2 without M3 collapses
($0.61$, grad-norm CoV explodes to $4.97$); the full recipe is stable ($0.91$).
Removing M2 or M3 from the full recipe collapses ($0.59$/$0.61$) while removing
M1 costs only $\sim$2 points but creates spikes in grad-norm. \texttt{val\_rep} $=$ mean 20-char
repetition rate; \texttt{grad CoV} $=$ coefficient of variation of
\texttt{actor/grad\_norm} over training.}
\label{tab:ablation_loo}
\centering
\small
\begin{tabular}{l c c c r r r}
\toprule
Configuration & M1 & M2 & M3 & M-500 (100) & val\_rep & grad CoV \\
\midrule
RS-SFT (positive-only)      & ---        & ---        & ---        & 0.71 & 0.55 & 0.34 \\
$+$ z-score (M1)            & \checkmark & ---        & ---        & 0.78 & 0.30 & 0.29 \\
$+$ negatives (M1$+$M2)     & \checkmark & \checkmark & ---        & 0.61 & 0.49 & 4.97 \\
$+$ clipping $=$ \opft{}    & \checkmark & \checkmark & \checkmark & 0.91 & 0.08 & 0.18 \\
\midrule
\opft{} $-$ negatives (M2)  & \checkmark & ---        & \checkmark & 0.59 & 0.57 & 0.33 \\
\opft{} $-$ z-score (M1)    & ---        & \checkmark & \checkmark & 0.89 & 0.06 & 0.46 \\
\bottomrule
\end{tabular}
\end{table}

\section{A second model family: Phi-4-mini-reasoning}
\label{app:phi4mini}

All experiments in the main text use Qwen3-4B. To check that the stability
gap is a property of the update rule and not of one model family, we repeat
the coding configuration on Phi-4-mini-reasoning \citep{xu2025phi4minireasoning}
--- a different pretraining corpus, tokenizer, and architecture --- holding
everything else fixed: DeepCoder verified 16K, \texttt{bs}$=256$ prompts,
$N=16$ rollouts at temperature 1.0, max response length 8192, lr $=10^{-6}$,
binary unit-test reward, 406 training steps ($\approx$7 epochs), single run,
evaluated on LCBv6 with greedy decoding.

\begin{figure}[h]
\centering
\includegraphics[width=0.62\columnwidth]{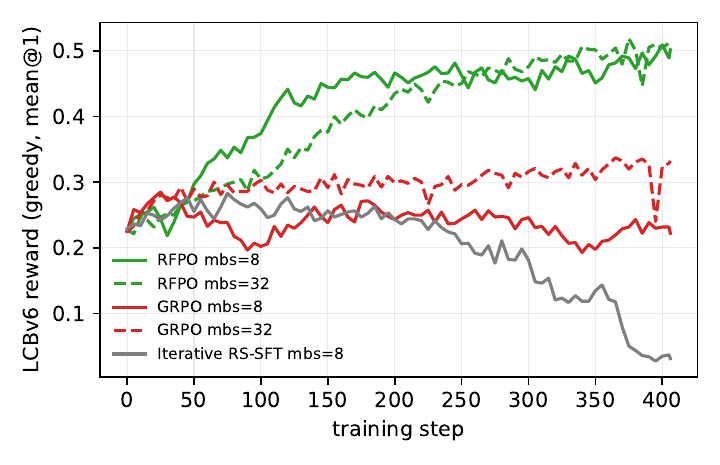}
\caption{Phi-4-mini-reasoning on DeepCoder-16K, LCBv6 validation (greedy).
\opft{} (green) is flat in \texttt{mbs}: the $8$ and $32$ curves coincide.
GRPO (red) is not: \texttt{mbs}$=32$ underperforms \opft{} by $\approx$18
points and \texttt{mbs}$=8$ decays back to its initialization. The
positive-only iterative RS-SFT baseline (gray) peaks at step 75 and then
collapses.}
\label{fig:phi4mini}
\end{figure}

\Cref{fig:phi4mini} and \cref{tab:phi4mini} show the same ordering as
\cref{sec:drift_mbs}. \opft{} is insensitive to \texttt{mbs}: at
\texttt{mbs}$=8$ ($K=32$) and \texttt{mbs}$=32$ ($K=8$) it rises smoothly from
$0.23$ to a late-mean of $0.494$ and $0.499$ --- the two curves are within
run-to-run noise of each other despite a $4\times$ difference in drift.
GRPO is not. At \texttt{mbs}$=32$ it climbs slowly to $0.314$, and at
\texttt{mbs}$=8$ it peaks at $0.285$ by step 25 and then \emph{loses} that
gain, ending at $0.229$ --- no better than the untrained model. The drift
diagnostics identify the mechanism: at \texttt{mbs}$=8$ GRPO's
$\pi_\theta$--$\pi_{\text{old}}$ KL spikes to $0.134$ (nearly $100\times$ its
own median of $1.7\times10^{-3}$) around steps 88--92 and 125--126, exactly
where the reward curve turns over, and its token entropy swings over
$[0.024, 1.09]$; at \texttt{mbs}$=32$ the same KL stays below $1.4\times10^{-3}$
throughout and entropy decays smoothly within $[0.19, 0.43]$.

The positive-only iterative RS-SFT baseline is unstable on this model too, and
in the characteristic way: it improves for 75 steps (to $0.283$), then
collapses to $0.034$. Its own diagnostics show the degenerate fixed point of
\cref{sec:method_mechanisms} directly --- mean response length falls from
$4892$ to $978$ tokens while the NLL of its \emph{own} selected samples
\emph{rises} from $0.42$ to $0.63$, i.e.\ the policy is concentrating on
progressively shorter outputs that it fits progressively worse. \opft{}'s
mean response length stays flat ($\approx$5000 tokens) over the same window.

\begin{table}[h]
\caption{Phi-4-mini-reasoning, LCBv6 (greedy, mean@1), single runs.
\emph{Peak} is the best evaluation and \emph{late} the mean over the final 25
steps; a large peak--late gap is the signature of instability. \opft{} has
none at either \texttt{mbs}; GRPO at \texttt{mbs}$=8$ gives back its entire
gain, and iterative RS-SFT collapses.}
\label{tab:phi4mini}
\centering
\begin{tabular}{llccc}
\toprule
Method & \texttt{mbs} ($K$) & init & peak (step) & late \\
\midrule
\opft{}            & 32 ($K$=8)  & 0.228 & 0.519 (375) & \textbf{0.499} \\
\opft{}            & 8 ($K$=32)  & 0.226 & 0.509 (400) & \textbf{0.494} \\
GRPO               & 32 ($K$=8)  & 0.224 & 0.337 (365) & 0.314 \\
GRPO               & 8 ($K$=32)  & 0.225 & 0.285 (25)  & 0.229 \\
Iterative RS-SFT   & 8 ($K$=32)  & 0.229 & 0.283 (75)  & 0.034 \\
\bottomrule
\end{tabular}
\end{table}

\section{Loss scaling does not affect training}
\label{app:scale2n}

\opft{}'s update sums a per-token loss over the $N$ rollouts of each prompt
group, a variable mix of positives and negatives. Whether the advantages carry
M1's $2/N$ factor --- equivalently, whether the per-group loss is normalized by
$N/2$ --- is an implementation choice: it rescales the gradient by
a constant factor that the optimizer's adaptive step size largely absorbs.
The loss is averaged over all response tokens in the mini-batch, masked tokens
included (their weight is zero but they count in the denominator), matching verl's
token-mean aggregation, so masking a token shrinks the update rather than
re-weighting the remaining tokens.
Relative to GRPO's advantages, \opft{}'s differ by a constant factor ($2/N$, and
$\sqrt{N/(N-1)} \approx 1.03$ at $N=16$ because \opft{} normalizes by the population and
verl by the sample standard deviation), which a scale-invariant optimizer absorbs. To
confirm that our results do not depend on this choice, we ran the coding
configuration (DeepCoder-16K, Qwen3-4B, \texttt{mbs}$=8$) twice, identical in
every logged hyperparameter except the scaling flag.

\begin{figure}[h]
\centering
\includegraphics[width=0.62\columnwidth]{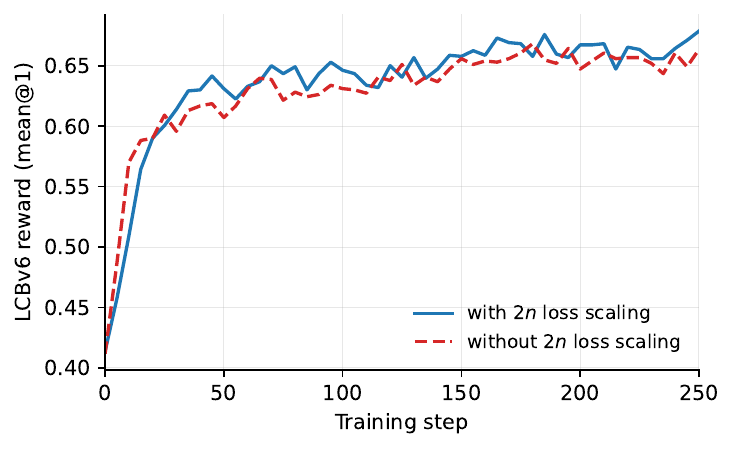}
\caption{Loss scaling does not affect training (LCBv6, DeepCoder-16K,
Qwen3-4B, \texttt{mbs}$=8$, first 250 steps). The two runs are identical in
every logged hyperparameter except whether the advantages carry the $2/N$
factor of M1; their curves coincide to within run-to-run noise (mean absolute difference
$0.013$ over 51 evaluation points).}
\label{fig:scale2n}
\end{figure}

\Cref{fig:scale2n} shows the two learning curves over the first 250 steps.
They are indistinguishable: both rise from $0.41$ at initialization to
$\approx0.66$, with a mean absolute difference of $0.013$ across the 51 logged
evaluation points --- the same order as the run-to-run variation reported in
\cref{tab:drift_mbs}. Over the final ten evaluations the two settings average
$0.664$ (with scaling) and $0.655$ (without). The rescaling is not a no-op at
the gradient level --- the median logged gradient norm differs by $7.6\times$
between the two runs ($2.2\times10^{-3}$ with scaling, $1.7\times10^{-2}$
without) --- so this is an empirical check on AdamW's scale invariance rather
than a restatement of it, which is worth making because $\epsilon$, decoupled
weight decay, and global gradient-norm clipping all break that invariance
exactly. We therefore report results
without distinguishing the two, and the choice is not one of \opft{}'s three
mechanisms.

\section{Gradient-estimator probe: protocol and values}
\label{app:gradprobe}

This appendix gives the protocol behind \cref{fig:gradprobe} (\cref{sec:method_biasvar})
and its values (\cref{tab:gradprobe}).

\paragraph{Checkpoints and inner loop.} We probe the step-25 checkpoints of two math
runs at \texttt{mbs}=8: an \opft{} run ($\delta = 0.1$), probed with \opft{}'s own
update, and a GRPO run ($\beta = 0.001$, reference $=$ initial model), probed with
GRPO's own update and the run's saved optimizer state. The \opft{} run's optimizer
state could not be restored, so its probe starts from a fresh AdamW (lr $10^{-6}$,
$\beta = (0.9, 0.999)$, weight decay 0.01, gradient clipping 1.0, as in training).
Both checkpoints are early in training; later checkpoints were not probed.

\paragraph{Rollout batch and updates.} From each checkpoint we draw one rollout batch
of 256 POLARIS prompts (prompts longer than 1024 tokens dropped), $N = 16$ rollouts at
temperature 1.0, at most 4096 tokens, sampled with vLLM and scored with the training
reward. The batch is split into 32 fixed mini-batches of 8 prompts; the inner loop
takes $k = 1, \dots, 31$ updates on them in order, and we record $\theta_k$ at
$k \in \{0, 1, 2, 4, 8, 16, 31\}$.

\paragraph{Estimators and normalization.} At each recorded $\theta_k$ we compute, on
every mini-batch $b$ of the rollout batch, $g^{\text{IS}}_b$ from verl's clipped
policy-gradient loss ($\epsilon = 0.2$, dual clip $c = 3$, no KL term) and
$g^{\text{RF}}_b$ from \opft{}'s masked weighted-MLE loss, with the same group-relative
advantages and the same token-mean normalization, so the comparison does not depend on
\opft{}'s constant advantage scaling (\cref{app:scale2n}). $\bar g$ is the mean over the 32 mini-batches.

\paragraph{Reference and noise correction.} The reference $g_{\text{ref}}$ is the
on-policy estimator at $\theta_k$, averaged over the 32 mini-batches of a \emph{fresh}
4096-sample batch drawn from $\theta_k$ on the same prompts. The noise variance
$\sigma^2$ of a batch mean is estimated by a half-split (even- and odd-indexed rollouts
within each group, each normalized by the full token count, $\sigma^2 = \|H_0 -
H_1\|^2/4$). Since the two batches are independent,
$\mathbb{E}\|\bar g - g_{\text{ref}}\|^2 = \|\mathrm{Bias}\|^2 + \sigma^2_{\bar g} +
\sigma^2_{\text{ref}}$, and we report $\|\bar g - g_{\text{ref}}\|^2 / (\sigma^2_{\bar g} +
\sigma^2_{\text{ref}})$, which is $\approx 1$ when there is no measurable bias (the
\emph{noise floor}). The MSE is $\frac{1}{32}\sum_b \|g_b - g_{\text{ref}}\|^2 -
\sigma^2_{\text{ref}}$, i.e.\ corrected for the reference's own noise.

\paragraph{Parameter subset.} Gradients are taken with respect to the last two decoder
layers and the final norm ($2.0\times10^{8}$ parameters, the same for every $k$ and
estimator), on a bf16 replica of the actor's weights, accumulated in fp32.

\paragraph{Limits.} There is one fresh reference draw per $k$, and the reference's
signal-to-noise ratio is about 1, so ratios of 1--2 are not resolvable; the values of
10--27 for $\hat g_{\text{IS}}$ are. The fixed 8-prompt mini-batches differ from the
training loop, which packs active samples. At the initial model, where drift stays
small, the effect is weak: with \opft{}'s update $\hat g_{\text{IS}}$ reaches 1.8--2.0
at $k = 31$ against 1.2--1.3 for $\hat g_{\text{RF}}$, and with GRPO's update neither
estimator shows a bias that grows with $k$.

\begin{table}[h]
\caption{Gradient-estimator probe values (\cref{fig:gradprobe}): squared bias relative
to the noise floor, and MSE ($\times10^{-3}$), per number of inner updates $k$.}
\label{tab:gradprobe}
\centering
\small
\begin{tabular}{r rrrr rrrr}
\toprule
 & \multicolumn{4}{c}{\opft{} checkpoint (step 25)} & \multicolumn{4}{c}{GRPO checkpoint (step 25)} \\
\cmidrule(lr){2-5} \cmidrule(lr){6-9}
 & \multicolumn{2}{c}{bias$^2$/noise} & \multicolumn{2}{c}{MSE} & \multicolumn{2}{c}{bias$^2$/noise} & \multicolumn{2}{c}{MSE} \\
$k$ & IS & RF & IS & RF & IS & RF & IS & RF \\
\midrule
0 & 1.1 & 1.1 & 3.44 & 3.44 & 1.0 & 1.0 & 2.45 & 2.45 \\
1 & 1.1 & 0.9 & 3.36 & 3.44 & 2.7 & 1.0 & 2.53 & 2.47 \\
2 & 1.3 & 1.0 & 3.38 & 3.52 & 2.7 & 1.2 & 2.51 & 2.50 \\
4 & 4.8 & 1.0 & 4.25 & 3.59 & 1.9 & 1.0 & 2.34 & 2.51 \\
8 & 15.5 & 1.1 & 6.71 & 3.73 & 2.1 & 1.1 & 2.31 & 2.49 \\
16 & 26.3 & 1.4 & 9.58 & 3.84 & 9.3 & 1.4 & 3.23 & 2.55 \\
31 & 27.5 & 1.7 & 9.60 & 3.82 & 10.3 & 3.9 & 3.27 & 2.60 \\
\bottomrule
\end{tabular}
\end{table}

\paragraph{Sequence-level drift.} On the same 4096 rollouts, after 31 inner updates
the mean absolute summed log-ratio per sequence, $|\sum_t
\log(\pi_\theta(y_t)/\pi_{\text{old}}(y_t))|$, is 3.3 at the initial model, 8.6 at the
\opft{} checkpoint and 11.7 at the GRPO checkpoint (99th percentile up to 81), while
the length-normalized log-ratio of a sequence stays small (99th percentile
$3.4\times10^{-3}$ to $1.0\times10^{-2}$): per-token ratios stay within about 1\% of 1. Per-token trust regions therefore do not
bound drift at the sequence level (\cref{app:future_work}).

\section{Per-run final evaluations}
\label{app:fullresults}

\Cref{tab:full_eval_runs} lists the individual math runs behind \cref{tab:full_eval},
including the two mixed cells of \cref{tab:twobytwo}, and \cref{tab:coding_runs} the
coding runs behind \cref{tab:twobytwo}. All math values are for the step-208 checkpoint,
greedy unless marked avg@16.

\paragraph{Re-runs behind \cref{tab:drift_mbs}.} The math cells of \cref{tab:drift_mbs}
come from re-runs of every math arm in that table with configurations identical to the
original runs (whose curves the figures show), except that the three runs of an arm set
the data-sampler seed to 42/43/44 (the original runs left it unset) and the GRPO runs,
which carry the KL term, use FSDP2 (\cref{app:details}). Each re-run is evaluated at step 208 through
the same validation code as \cref{tab:full_eval}; \cref{tab:retrain_runs} lists the
per-run values. The re-runs agree with the independent runs of \cref{tab:full_eval}:
\opft{} at \texttt{mbs}=8 reaches 0.875 against 0.869 there, and GRPO 0.824 against 0.823.

\begin{table}[h]
\caption{Re-runs behind the math cells of \cref{tab:drift_mbs}, per run (step 208;
runs 1--3 use data seeds 42/43/44). MATH-500 greedy, AIME avg@16. $K$: optimizer steps
per batch. $^\dagger$This run's AIME evaluation did not complete, so AIME for this arm has
$n=2$.}
\label{tab:retrain_runs}
\centering
\small
\setlength{\tabcolsep}{4pt}
\begin{tabular}{l c c c c c}
\toprule
Arm & $K$ & run & MATH-500 & AIME-24 & AIME-25 \\
\midrule
GRPO & 1  & 1 & 0.878 & 0.421 & 0.346 \\
     &    & 2 & 0.884 & 0.442 & 0.327 \\
     &    & 3 & 0.874 & 0.398 & 0.338 \\
GRPO & 8  & 1 & 0.846 & 0.346 & 0.308 \\
     &    & 2 & 0.876 & 0.367 & 0.310 \\
     &    & 3 & 0.862 & 0.346 & 0.306 \\
GRPO & 32 & 1 & 0.824 & 0.285 & 0.233 \\
\opft{} & 8  & 1 & 0.886 & 0.423 & 0.354 \\
        &    & 2 & 0.862 & 0.425 & 0.348 \\
        &    & 3 & 0.872 & 0.425 & 0.350 \\
\opft{} & 32 & 1 & 0.878 & 0.435 & 0.356 \\
        &    & 2 & 0.878 & 0.400 & 0.338 \\
        &    & 3 & 0.868 & 0.388 & 0.342 \\
\midrule
GRPO, bypass & 8  & 1 & 0.718 & 0.133 & 0.054 \\
GRPO, bypass & 32 & 1 & 0.786 & 0.198 & 0.196 \\
\opft{}, bypass & 8  & 1 & 0.868 & 0.408 & 0.331 \\
                &    & 2 & 0.860 & 0.410 & 0.329 \\
                &    & 3 & 0.870 & n/a$^\dagger$ & n/a$^\dagger$ \\
\opft{}, bypass & 32 & 1 & 0.794 & 0.285 & 0.198 \\
\bottomrule
\end{tabular}
\end{table}
\begin{table}[h]
\caption{Math, per run (step 208; \texttt{mbs}=8 unless noted).}
\label{tab:full_eval_runs}
\centering
\small
\setlength{\tabcolsep}{4pt}
\begin{tabular}{l c cc cc}
\toprule
 & & \multicolumn{2}{c}{AIME 2024} & \multicolumn{2}{c}{AIME 2025} \\
\cmidrule(lr){3-4} \cmidrule(lr){5-6}
Run & MATH-500 & greedy & avg@16 & greedy & avg@16 \\
\midrule
Base model & 0.678 & 0.200 & 0.117 & 0.167 & 0.094 \\
GRPO, \texttt{mbs}=256, $\beta$=0.001 & 0.874 & 0.500 & 0.417 & 0.333 & 0.340 \\
GRPO, $\beta$=0.001, run 1 & 0.836 & 0.300 & 0.310 & 0.233 & 0.235 \\
GRPO, $\beta$=0.001, run 2 & 0.812 & 0.333 & 0.317 & 0.233 & 0.237 \\
GRPO, $\beta$=0.001, run 3 & 0.822 & 0.333 & 0.279 & 0.200 & 0.223 \\
GRPO, $\beta$=0 (ratio $+$ ratio clip) & 0.818 & 0.367 & 0.302 & 0.267 & 0.244 \\
Ratio $+$ absolute clip, $\beta$=0 & 0.860 & 0.267 & 0.308 & 0.200 & 0.235 \\
Ratio-free $+$ ratio clip & 0.858 & 0.333 & 0.369 & 0.367 & 0.317 \\
GSPO & 0.702 & 0.067 & 0.102 & 0.100 & 0.085 \\
DAPO (no dynamic sampling) & 0.858 & 0.233 & 0.323 & 0.267 & 0.298 \\
\opft{}, $\delta$=0.05 & 0.874 & 0.467 & 0.404 & 0.433 & 0.313 \\
\opft{}, $\delta$=0.2 & 0.858 & 0.367 & 0.342 & 0.300 & 0.310 \\
\opft{}, $\delta$=0.1, run 1 & 0.882 & 0.400 & 0.396 & 0.367 & 0.356 \\
\opft{}, $\delta$=0.1, run 2 & 0.866 & 0.333 & 0.381 & 0.367 & 0.310 \\
\opft{}, $\delta$=0.1, run 3 & 0.860 & 0.400 & 0.410 & 0.400 & 0.367 \\
\bottomrule
\end{tabular}
\end{table}

\begin{table}[h]
\caption{Coding (LCBv6, Qwen3-4B, DeepCoder-16K, \texttt{mbs}=8, one node per run, at
most 300 steps): median and mean over steps 240--270. The ratio-free runs have isolated
single-point evaluation failures (\opft{} run 1: 0.00 at step 240 and 0.41 at 265;
\opft{} run 2: 0.42 at 245; each followed by 0.66--0.68 at the next evaluation), most
likely validations at job restarts after preemption; the median is robust to them,
the mean is not.}
\label{tab:coding_runs}
\centering
\small
\begin{tabular}{l c c c}
\toprule
Run & median & mean & note \\
\midrule
\opft{}, run 1 & 0.67 & 0.54 & \\
\opft{}, run 2 & 0.66 & 0.63 & \\
Ratio-free $+$ ratio clip & 0.65 & 0.65 & \\
GSPO & 0.27 & 0.26 & \\
Ratio $+$ absolute clip & 0.00 & 0.00 & collapsed \\
GRPO, $\beta$=0.001, run 1 & --- & --- & collapsed to 0.12 by step 90, stopped \\
GRPO, $\beta$=0.001, run 2 & --- & --- & collapsed to 0.23 by step 110, stopped \\
DAPO (no dynamic sampling) & --- & --- & collapsed to 0.09 by step 90, stopped \\
\bottomrule
\end{tabular}
\end{table}

\end{document}